\documentclass{bmvc2k}

\title{Adversarially Robust Few-Shot Anomaly Detection with Vision Foundation Models}

\addauthor{Akib Mohammed Khan}{ak9029@rit.edu}{1}
\addauthor{Bartosz Krawczyk}{bartosz.krawczyk@rit.edu}{1}

\addinstitution{
Center for Imaging Science\\
 Rochester Institute
 of Technology\\
 Rochester, NY, USA
}

\runninghead{Khan and Krawczyk}{Adversarially Robust FSAD with VFMs}

\usepackage{bm}
\usepackage{amssymb}
\usepackage{algorithm}
\usepackage{algpseudocode}
\usepackage{amsmath}
\usepackage{amsthm}
\usepackage{booktabs, multirow}
\usepackage{makecell}
\usepackage{xcolor}

\usepackage{subfigure}
\usepackage{hyperref}
\begin{document}

\maketitle

\begin{abstract}
Vision foundation models such as DINOv2 enable strong few-shot anomaly
detection (FSAD) through simple non-parametric $k$-nearest-neighbor
($k$-NN) scoring over frozen patch features. Existing robust anomaly
detection methods assume access to large normal-class training sets and
adversarial training of the feature extractor. The few-shot regime, where
the detector consists of a frozen backbone and a non-parametric scorer,
has remained unaddressed despite its widespread practical deployment.
This study addresses the gap by developing a white-box attack framework
adapted to $k$-NN memory-bank detectors and introduces an adversarially
robust FSAD method which is training-free with respect to the backbone. We introduce the DistanceProbe (DP), a lightweight MLP trained by regression to predict per-patch $k$-NN distance directly from frozen features, providing the differentiable
proxy required for gradient-based attacks against non-parametric scorers.
We further propose a two-level combined defense that operates using only
the normal reference support set images. At the input level,
PatchShift (PS) applies random pixel shifts to exploit the
misalignment of adversarial perturbations with the encoder's
patch grid, aggregating scores across multiple shifted views via
element-wise median pooling. At the feature level we employ the
FeaturePurifier (FP), a lightweight residual MLP trained on clean
and adversarially augmented support features to project perturbed
representations back toward the clean feature manifold. Through comprehensive experiments under various adversarial settings, we show that our method exhibits robust detection and localization with performance gains of $\sim27\%$ image-level AUROC and $\sim51\%$ pixel-level PRO over the undefended attacked baseline across MVTec-AD, VisA, and MPDD at $k=4$ shots, while preserving clean accuracy within $3\%$, matching full-shot adversarially-trained baselines on pixel-level AUROC, and holding under adaptive attack. The code is available at \href{https://github.com/akibkhan1/robust-fsad}{https://github.com/akibkhan1/robust-fsad}.
\end{abstract}

\section{Introduction}
\label{sec:introduction}

Visual Anomaly Detection (AD) and Anomaly Localization (AL) are widely used techniques in various domains including manufacturing, medical imaging and cyber-security \cite{bergmann2019mvtec, liu2024deep, fernando2021deep, tian2023self, siddiqui2019detecting} where unannotated defects must be detected
and localized from nominal exemplars only \cite{bergmann2022beyond, li2021cutpaste, zou2022spot}, to predict an image or pixel to be normal or anomalous. With a focus on industrial images, existing deep learning (DL) based AD methods perform exceptionally when normal samples are abundant \cite{bergmann2019mvtec, defard2021padim, roth2022towards, mousakhan2024anomaly} with recent work emphasizing
deployment efficiency~\cite{batzner2024efficientad} and patch-level
consistency~\cite{hyun2024reconpatch}. A practical bottleneck remains
at the cost of curating large nominal training sets,
which has driven interest in few-shot anomaly detection
(FSAD)~\cite{li2024musc, zhou2024anomalyclip, cao2024adaclip, huang2022registration, fang2023fastrecon, jeong2023winclip, zhu2024toward, gu2025univad, xie2023pushing, damm2025anomalydino}. Within this regime, vision foundation
models such as DINOv2~\cite{oquab2023dinov2} have proven to be exceptionally
effective with AnomalyDINO~\cite{damm2025anomalydino} demonstrating that a
frozen DINOv2 backbone combined with non-parametric $k$-nearest-neighbor
($k$-NN) scoring over a patch-level memory bank attains
state-of-the-art FSAD performance without any detector training.

\begin{figure}[t]
\centering
\subfigure{\includegraphics[width=0.46\linewidth, trim={10pt 20pt 1pt 1pt}, clip]{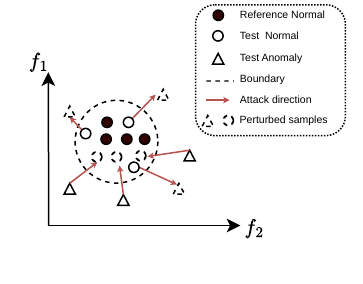}}
\hfill
\subfigure{\includegraphics[width=0.51\linewidth]{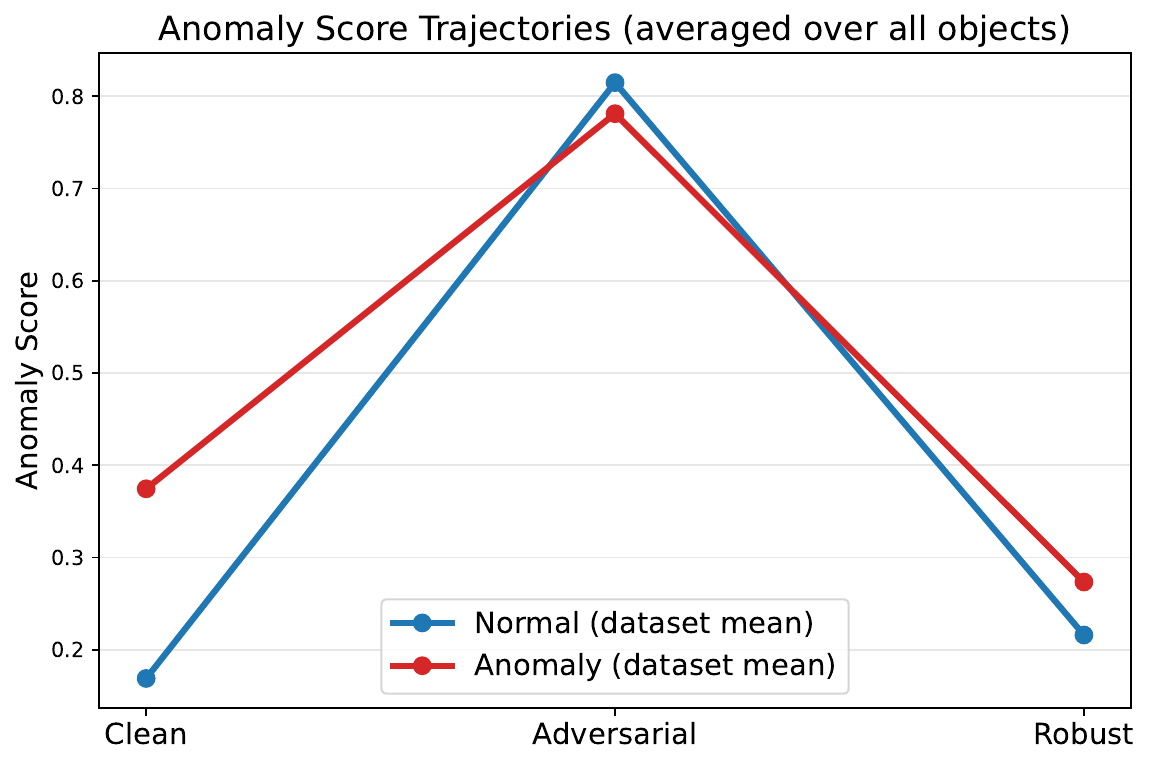}}
\caption{\emph{Left.}~Feature space overview where a gradient-based attack pushes normal patches across the boundary into anomalous regions and drags anomalous patches toward the reference-normal cluster, inverting the $k$-NN ranking.
\emph{Right.}~On MVTec-AD, clean inputs separate normal (blue) and anomalous (red) scores correctly; under PGD-20 at $\epsilon=8/255$ the ordering collapses and flips; our proposed defense restores it.}
\label{fig:teaser}
\end{figure}

This frozen-backbone, training-free design is attractive for
deployment but raises a question that has gone unexamined: \emph{are such
detectors robust to adversarial perturbations?} Adversarial examples
remain a primary threat to deep vision
systems~\cite{goodfellow2014explaining, madry2017towards, akhtar2018threat}, and recent
work has shown that anomaly detectors built on trained backbones are
similarly brittle~\cite{goodge2021robustness, mirzaei2025adversarially,
nafez2025patchguard, cao2024adversarially}. Existing robust AD and AL
methods, however, all share two assumptions incompatible with the
FSAD paradigm: they require (i)~access to a large nominal training
set and (ii)~adversarial training of the feature extractor itself, or
of a large generative backbone~\cite{cao2024adversarially}. Neither
assumption holds when the detector consists of a frozen foundation
model and $k$ support images. The adversarial robustness of few-shot,
foundation-model-based anomaly detectors has therefore
remained an open question, despite the practical prevalence of this
deployment pattern.

Addressing this gap is non-trivial for two reasons. First, evaluating
robustness requires a strong gradient-based attack, but the
non-parametric $k$-NN scorer is only piecewise differentiable through
its active minimizer, and the identity of the nearest neighbor is
itself unstable under perturbation, yielding a gradient signal too
noisy to drive standard gradient-based attacks. Second,
defending the pipeline cannot rely on retraining the backbone or
expanding the memory bank, since both would violate the few-shot,
frozen-backbone constraint that motivates the setting.

To address these challenges we introduce the \emph{DistanceProbe}
(DP), a lightweight MLP trained by regression to predict per-patch
$k$-NN distances directly from frozen DINOv2 features, providing the
differentiable surrogate required to mount gradient-based attacks against
non-parametric scorers. We subsequently propose a robust AD and AL method for the FSAD setting with a two-level defense pipeline that operates using only the $k$ support images. At the input level, \emph{PatchShift} (PS) exploits the observation
of Gu~\textit{et al.}~\cite{gu2022vision} that adversarial
perturbations are tightly coupled to the Vision Transformer's (ViT) patch
grid: small pixel shifts decorrelate the perturbation from the
tokenizer, and median-pooling scores across shifted views attenuates
its effect. At the feature level, the \emph{FeaturePurifier} (FP),
adapted from Dong~\textit{et al.}~\cite{dong2022improving}, learns a
residual MLP that projects perturbed features back toward the clean
manifold, trained entirely on the support set with both adversarial
and Gaussian-noise augmentations. Figure~\ref{fig:teaser} illustrates the adversarial vulnerability of FSAD and the robustness of our proposed method.
In feature space, a gradient-based attack against the $k$-NN scorer
inverts the normal-versus-anomalous distance ranking that the
detector relies on, and on MVTec-AD benchmark dataset \cite{bergmann2019mvtec} this
manifests as a complete collapse, and reversal of image-level
score separation, which our proposed defense recovers.

The framework is agnostic to the choice of foundation model. We adopt
DINOv2 following AnomalyDINO~\cite{damm2025anomalydino} as it is
purely vision-based, isolating the visual representation from
text-prompt confounds present in multi-modal
detectors~\cite{jeong2023winclip, zhou2024anomalyclip, cao2024adaclip}, and matching
the practical setting in which human annotators specify normality
through visual exemplars alone, without anomaly type descriptions or
auxiliary text prompts.

\textbf{Contributions.}
\begin{itemize}
\item We formalize adversarial robustness for few-shot anomaly
      detection with frozen vision foundation models, a setting
      unaddressed by prior work.
\item We introduce the \emph{DistanceProbe}, a regression-trained MLP
      surrogate that makes gradient based attacks tractable against
      non-parametric $k$-NN memory-bank scorers.
\item We propose \emph{PatchShift + FeaturePurifier}, a two-level
      defense that requires no backbone training, operates on only
      the $k$ support images, and is backbone-agnostic. 
\item Across MVTec-AD \cite{bergmann2019mvtec}, VisA \cite{zou2022spot}, and MPDD \cite{jezek2021deep}, our method recovers
      substantial robustness under strong gradient-based attacks,
      preserves clean performance within $\sim$3 points, matches or
      exceeds full-shot adversarially-trained baselines on AL, and holds under adaptive attacks.
\end{itemize}

\section{Related Works}
\label{sec:related_works}

\noindent\textbf{Industrial Anomaly Detection (IAD).} Anomaly
detection (AD) seeks to identify test samples that deviate from a
predefined notion of normality~\cite{ruff2021unifying, yang2024generalized}.
We focus on low-level sensory anomalies in industrial imagery, where
the dominant practical paradigm has converged on \emph{deep nearest
neighbor} AD: test patches are scored by their distance to the
closest entry in a memory bank of nominal patch features. This
approach traces back to early image-level
formulations~\cite{eskin2002geometric, bergman2020deep} and was
popularized at the patch level by Patch
SVDD~\cite{yi2020patch}, which trains a patch feature extractor
via self-supervision, and
SPADE~\cite{cohen2020sub}, which introduced multi-scale
feature-pyramid correspondences. PatchCore~\cite{roth2022towards}
later refined this paradigm using pretrained ImageNet classifiers
with coreset subsampling, establishing the de facto reference for
patch-level nearest-neighbor scoring. Subsequent variants include
graph-based modeling (GraphCore~\cite{xie2023pushing}), and batched zero-shot
extensions (MuSc~\cite{li2024musc}); orthogonal directions employ
the Mahalanobis distance~\cite{defard2021padim} or reverse
distillation~\cite{deng2022anomaly, tien2023revisiting}. Performance crucially
depends on the feature extractor; classical choices include
ResNets~\cite{he2016deep}, WideResNets~\cite{zagoruyko2016wide},
and ViTs~\cite{dosovitskiy2020image}, with
pretrained foundation models now setting the
state of the art~\cite{ren2025foundation, ammar2025foundation}.

\noindent\textbf{Foundation Models and Vision Backbones for AD.}
Self-supervised and vision-language foundation models have become
standard backbones for AD scoring. DINO~\cite{caron2021emerging} and
DINOv2~\cite{oquab2023dinov2} provide strong dense per-patch
representations via self-distillation, MAE~\cite{he2022masked} offers
reconstruction-pretrained features, and CLIP~\cite{radford2021learning}
contributes joint vision-language alignment that has powered most
recent zero-shot and few-shot AD
work~\cite{jeong2023winclip, zhou2024anomalyclip, zhu2024toward}.
Recent surveys~\cite{ren2025foundation, ammar2025foundation}
catalogue the rapid uptake of these backbones for AD across
industrial, medical, and video domains. While prior work has
extensively explored which backbone yields the best clean accuracy,
the question of how robust these backbones' frozen features are
under adversarial perturbation has remained largely unaddressed.

\noindent\textbf{Few-Shot Anomaly Detection.} FSAD targets rapid
adaptation with only a handful of nominal exemplars per category.
RegAD~\cite{huang2022registration} pioneered category-agnostic
alignment for few-shot transfer, and
FastRecon~\cite{fang2023fastrecon} proposed fast feature
reconstruction for cross-product generalization.
GraphCore~\cite{xie2023pushing} introduced graph-based feature
aggregation for the few-shot regime. Vision-language methods such
as WinCLIP~\cite{jeong2023winclip} adapted CLIP for zero- and
few-shot detection through compositional prompts;
AnomalyCLIP~\cite{zhou2024anomalyclip} learned object-agnostic
prompts for the zero-shot regime; and AdaCLIP~\cite{cao2024adaclip}
introduced hybrid learnable prompts.
InCTRL~\cite{zhu2024toward} cast FSAD as in-context residual
learning between query and few-shot prototype features.
MuSc~\cite{li2024musc} explored unsupervised mutual-scoring for the
zero-shot setting, while UniVAD~\cite{gu2025univad} proposed a
training-free unified framework combining component-aware patch
matching with graph-based logical reasoning. Closest to our setting,
AnomalyDINO~\cite{damm2025anomalydino} demonstrated that DINOv2
patch features paired with simple $k$-NN scoring achieve
state-of-the-art FSAD performance with no detector training. Our
work builds directly on this line and is the first to evaluate and
address \emph{adversarial robustness} within it.

\noindent\textbf{Adversarial Robustness of OOD and Anomaly
Detectors.} Adversarial examples~\cite{goodfellow2014explaining, madry2017towards}
remain a primary threat model for deep classifiers, and recent work
shows that out-of-distribution detectors are similarly
brittle~\cite{fort2021exploring, meinke2022provably, bitterwolf2020certifiably}.
AROS~\cite{mirzaei2025lyapunov} addressed adversarial OOD detection via
neural ODEs with Lyapunov-stabilized embeddings, exposing detectors
to auxiliary OOD samples during adversarial training. In anomaly
detection specifically, early work focused on autoencoder-based
detectors: APAE~\cite{goodge2021robustness} demonstrated reconstruction
attacks against autoencoder anomaly detectors on cyber-physical
sensor data and proposed an approximate-projection defense.
Subsequent work has targeted image-domain industrial AD with
adversarial training of the feature extractor:
PatchGuard~\cite{nafez2025patchguard} trained a ViT detector
adversarially with a patch-level mask-aware objective,
COBRA~\cite{mirzaei2025adversarially} adversarially trained a
ResNet-18 detector, and AdvRAD~\cite{cao2024adversarially} leveraged
a diffusion model's denoising property as an inherent robustness
mechanism. All three approaches assume the full-shot regime and
either train the backbone from scratch (PatchGuard, COBRA) or train
a large generative model on nominal data (AdvRAD); none operates
with a frozen pretrained foundation model or in the few-shot
regime. To our knowledge, no prior work addresses adversarial robustness of few-shot anomaly detection with frozen vision foundation models. Existing robust AD and AL
methods assume full-shot training data and adversarially trainable
backbones, both incompatible with the FSAD paradigm. We close this
gap by introducing the first attack framework targeting
non-parametric $k$-NN memory-bank detectors and a defense pipeline
that operates only on the $k$ support images, achieving robustness
without any backbone training.

\section{Preliminaries}
\subsection{DINOv2-Based Few-Shot Anomaly Detection}
\label{sec:prelim}

We follow the patch-level deep nearest-neighbor paradigm popularized by
PatchCore~\cite{roth2022towards} and recently shown to be highly
effective with DINOv2 features in the few-shot regime by
AnomalyDINO~\cite{damm2025anomalydino}. Let $x \in [0,1]^{3 \times H \times W}$ denote an RGB image. A frozen
DINOv2~\cite{oquab2023dinov2} backbone $f_\theta$ maps $x$ to a
sequence of $N$ patch tokens,
$f_\theta(x) = (p_1, \dots, p_N) \in \mathbb{R}^{N \times D}$, each represented by a $D$-dimensional feature embedding. Given $k$ nominal
reference images $X_{\mathrm{ref}} = \{x^{(i)}\}_{i=1}^{k}$, all
their patch tokens are collected into a memory bank
\begin{equation}
\label{eq:memory_bank}
\mathcal{M} \;=\; \bigcup_{i=1}^{k} \bigl\{ p_j^{(i)} \;:\; f_\theta(x^{(i)}) = (p_1^{(i)},\dots,p_N^{(i)}) \bigr\}.
\end{equation}
A test image $\bm{x}_{\mathrm{test}}$ is scored patch-by-patch against
$\mathcal{M}$. For each query patch $p$ the distance to
its nearest reference patch is computed,
\begin{equation}
\label{eq:nn_dist}
d_{\mathrm{NN}}(p;\, \mathcal{M}) \;=\; \min_{p_{\mathrm{ref}} \in \mathcal{M}} d\!\left(p,\, p_{\mathrm{ref}}\right),
\end{equation} where
\begin{equation}
\label{eq:cosine_equiv}
d(u, v) \;\triangleq\; \tfrac{1}{2} \bigl\lVert \tilde{u} - \tilde{v} \bigr\rVert_2^2 \;=\; 1 - \langle \tilde{u}, \tilde{v} \rangle \;=\; d_{\cos}(u, v),
\end{equation}
yielding an anomaly map $a(x_{\mathrm{test}}) \in \mathbb{R}^{N}$
with entries $a_j = d_{\mathrm{NN}}(p_j;\, \mathcal{M})$. For unit-norm
$\tilde{u}$ and $\tilde{v}$, $d_{\mathrm{NN}}$ is equivalent to
cosine distance. The
image-level score is the mean of the $1\%$ largest patch distances,
\begin{equation}
\label{eq:mean_top1p}
s(x_{\mathrm{test}}) \;=\; q\bigl(a(x_{\mathrm{test}})\bigr) \;\triangleq\; \mathrm{mean}\!\left(H_{0.01}\!\left(a(x_{\mathrm{test}})\right)\right),
\end{equation}
where $H_{0.01}(\cdot)$ returns the top-$1\%$ entries of its argument.
This statistic is the empirical tail-value-at-risk at the $99\%$ quantile
and follows AnomalyDINO~\cite{damm2025anomalydino}.


\subsection{Adversarial Robustness of Few-Shot Anomaly Detectors}
\label{sec:adv_fsad}
Consider a white-box, $\ell_p$-bounded adversary with access to
the backbone $f_\theta$, the score function $s$, and the
ground-truth $y$ of the test image. Given a test image $x$, the
adversary seeks an additive perturbation $\delta$ inside an
$\ell_p$ ball of radius $\epsilon$ that drives the per-patch
detector toward the opposite of the ground-truth ranking \cite{xu2020adversarial, yuan2019adversarial}:
\begin{equation}
\label{eq:threat_model}
\delta^\star \;=\; \arg\max_{\lVert \delta \rVert_p \leq \epsilon}\;\mathcal{L}(x + \delta,\, y), \qquad x + \delta \in [0,1]^{3 \times H \times W},
\end{equation}
This produces an adversarial sample $x^\star$, where $x^\star = x+\delta^\star$. 
A commonly used and effective adversarial attack method is the Projected Gradient Descent (PGD) ~\cite{madry2017towards} that maximizes the loss function iteratively by pushing towards the direction of the gradient sign of $\mathcal{L}(x^\star, y)$ with a defined step size $\alpha$. With respect to FSAD, the PGD objective is adapted to increase the anomaly score function $s(x,f)$ for normal samples and decrease it for anomalous samples \cite{nafez2025patchguard, mirzaei2025adversarially}. Specifically, starting from $x^{\star}_{0} = x$, each iteration takes a sign-gradient step in
pixel space and projects back onto the admissible $\ell_p$ ball:

\begin{equation}
\label{eq:pgd_step}
x^{\star}_{t+1} \;=\; \Pi_{\mathcal{B}_p(x,\,\epsilon)} \Bigl(x^{\star}_t + y \cdot \alpha \cdot \mathrm{sign}\!\left(\nabla_{x} s(x^{\star}_t; f_\theta)\right)\Bigr),
\end{equation}
where $y=+1$ for normal samples, $y=-1$ for anomalous samples, and $\mathcal{B}_p(x,\,\epsilon) = \{x' \in [0,1]^{3 \times H \times W} : \lVert x' - x \rVert_p \leq \epsilon\}$. The same procedure is applied to other attacks in our study.

\section{Methodology}
\label{sec:method}

\begin{figure}[t]
    \centering
    \includegraphics[width=\linewidth]{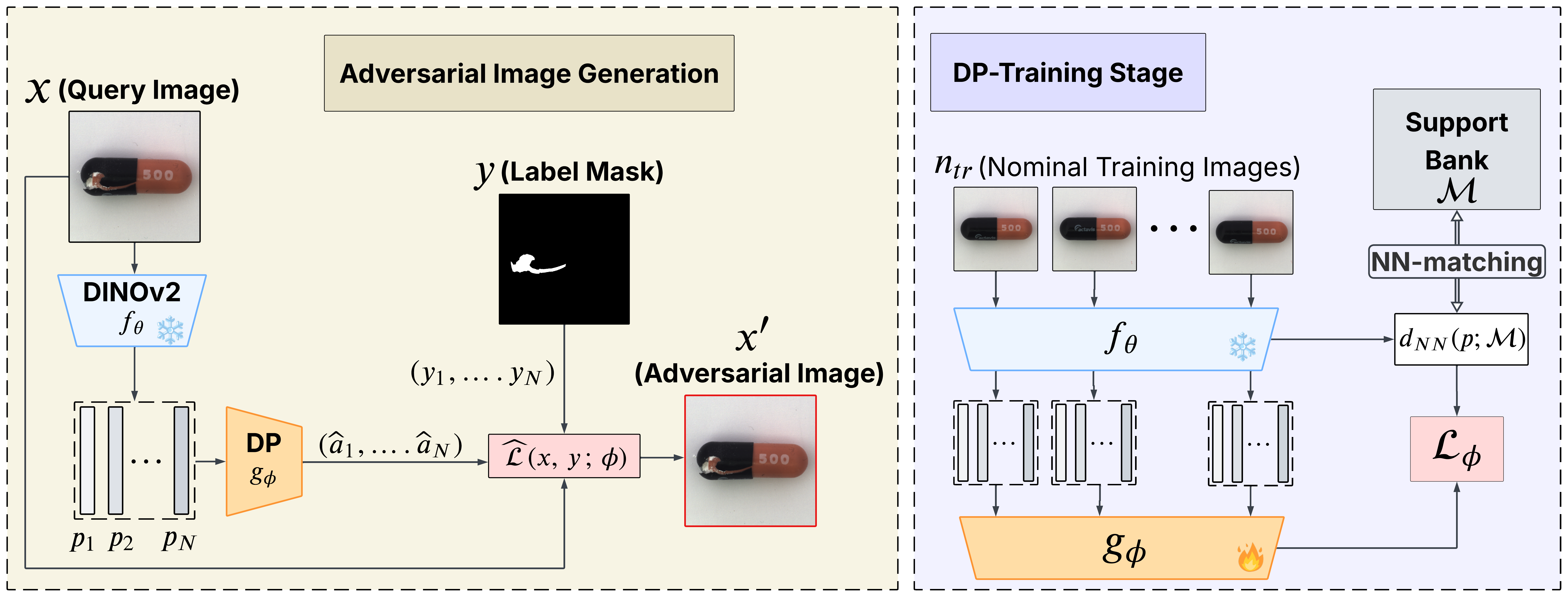}
    \caption{Attack framework. \emph{Left:}
    the frozen DINOv2 backbone $f_\theta$ extracts patch tokens
    $(p_1, \dots, p_N)$ from a query image $x$; the DP $g_\phi$
    predicts per-patch anomaly scores $(\widehat{a}_1, \dots, \widehat{a}_N)$
    from these tokens, and the mask-aware surrogate loss
    $\widehat{\mathcal{L}}(x, y; \phi)$ provides the gradient signal that
    PGD uses to craft the adversarial image $x'$. \emph{Right:} the probe $g_\phi$ is trained beforehand by regression to
    predict the per-patch nearest-neighbor distance $d_{\mathrm{NN}}(p;
    \mathcal{M})$ between a frozen-backbone feature $p$ and its closest
    match in the support memory bank $\mathcal{M}$, computed across
    $n_{\mathrm{tr}}$ nominal training images. Only $g_\phi$ is trained;
    $f_\theta$ remains frozen throughout.}
\label{fig:attack_framework}
\end{figure}

\subsection{Attack Framework: A Learned Surrogate for Non-Parametric Scoring}
\label{sec:attack}
Prior adversarial attacks on parametric anomaly detectors use the
detector's own head as the source of gradients. In the absence of such
a head for memory-bank methods with a frozen backbone, we propose an auxiliary probe called the \emph{DistanceProbe (DP)} on the support set features extracted from DINOv2, designed to match the $k$-NN scorer it must approximate (Figure~\ref{fig:attack_framework}). \emph{DP} is a three-layer multi-layer perceptron (MLP)
$g_\phi: \mathbb{R}^D \to \mathbb{R}$ trained \emph{by regression} to
predict the per-patch nearest-neighbor distance to the support memory
bank from a normalized patch token,
\begin{equation}
\label{eq:probe_arch}
g_\phi(\tilde{p}) \;=\; W_3\,\sigma\!\bigl(W_2\, \sigma(W_1 \tilde{p} + b_1) + b_2\bigr) + b_3,
\end{equation}
where $\sigma(\cdot) = \max(\cdot,0)$ is the ReLU activation,
$W_1 \in \mathbb{R}^{h \times D}$, $W_2 \in \mathbb{R}^{h \times h}$,
$W_3 \in \mathbb{R}^{1 \times h}$, and $\tilde{p}$ denotes an
$\ell_2$-normalized patch token.

The probe is trained on patch tokens extracted from $n_{\mathrm{tr}}$
nominal training images that are disjoint from the few-shot support set
$X_{\mathrm{ref}}$, using the same backbone $f_\theta$ and preprocessing
pipeline as the detector. For each such training image we extract
$M_{tr}$ masked patch tokens and label each token with its
nearest-neighbor distance to the support bank $\mathcal{M}$ under the
metric of Eq.~\eqref{eq:cosine_equiv}. To stabilize training across the
wide dynamic range of $d_{\mathrm{NN}}$ values, we regress on the
log-transformed target $\log(1 + d_{\mathrm{NN}})$. Letting
$\mathcal{D}_{\mathrm{tr}} = \{(\tilde{p}_i, y_i)\}_{i=1}^{M_{\mathrm{tr}}}$
with $y_i = \log\!\bigl(1 + d_{\mathrm{NN}}(p_i;\,\mathcal{M})\bigr)$,
the probe is fit by minimizing
\begin{equation}
\label{eq:probe_loss}
\phi^\star \;=\; \arg\min_\phi\; \frac{1}{M_{\mathrm{tr}}}\sum_{i=1}^{M_{\mathrm{tr}}}\!\bigl(g_\phi(\tilde{p}_i) - y_i\bigr)^2.
\end{equation}
The trained probe is then frozen and used only
at attack time. Training-data access for the attacker is consistent with
the white-box threat model: the attacker may use additional nominal
images outside the few-shot support to build the surrogate, but the
defender still operates with only $k$ support images at inference.

With the \emph{DP}, the per-patch score for the attacker becomes
\begin{equation}
\label{eq:probe_score}
\widehat{a}_j(x) \;=\; g_\phi\!\bigl(\widetilde{f_\theta(x)}_j\bigr), \qquad j = 1, \dots, N,
\end{equation}
which is end-to-end differentiable in $x$ through the frozen
backbone. Given the ground truth pixel map of a test image, the surrogate adversarial loss $\widehat{\mathcal{L}}$ is then the mask-aware difference between the
mean per-patch score in nominal regions and the mean per-patch score in anomalous regions:
\begin{equation}
\label{eq:adv_loss_probe}
\widehat{\mathcal{L}}(x,\,y\,;\,\phi) \;=\; \frac{1}{|\mathcal{I}_0|}\sum_{j \in \mathcal{I}_0}\!\widehat{a}_j(x) \;-\; \frac{1}{|\mathcal{I}_1|}\sum_{j \in \mathcal{I}_1}\!\widehat{a}_j(x),
\end{equation}
where $\mathcal{I}_0 = \{j : y_j = 0\}$ and $\mathcal{I}_1 = \{j : y_j = 1\}$, and $\widehat{a}_j(x)$ is the per-patch score at patch $j$. Maximizing $\widehat{\mathcal{L}}$ simultaneously inflates the score on nominal patches and suppresses it on anomalous ones. 
The PGD iteration of Eq.~\eqref{eq:pgd_step} is run with
the mean per-patch anomaly scores from Eq.~\eqref{eq:adv_loss_probe}. Because the probe is
calibrated against the true scorer's output by Eq.~\eqref{eq:probe_loss},
the gradient signal $\nabla_{\bm{x}}\widehat{\mathcal{L}}$ provides a
local linear approximation of the true non-parametric score that is
faithful in the neighborhood of clean patches, where the probe was
trained. The PGD update operates in pixel space; the
DINOv2 normalization $x \mapsto (x - \mu) / \sigma$
is composed into $f_\theta$, and per-channel scaling of $\alpha$ and
$\epsilon$ by $1/\sigma$ ensures the constraint is enforced in the
correct space.

\begin{figure}[t]
    \centering
    \includegraphics[width=\linewidth]{images/def_final.pdf}
    \caption{Defense framework. \emph{Left:}
    an adversarial image $x^{adv}$ is processed through both an undefended
    pipeline (top branch, feeding the frozen DINOv2 $f_\theta$ and
    NN-matching against the support bank $\mathcal{M}$, yielding a
    ``perturbed'' anomaly map and image-level anomaly score) and the proposed
    robust pipeline (bottom branch). PS produces $K$
    shift-then-cropped views $\{\Delta^{(1)}, \dots, \Delta^{(K)}\}$, each
    encoded by $f_\theta$ and refined by the FP $h_\psi$
    before NN-matching. Median aggregation across the $K$ per-shift
    distance maps yields the recovered anomaly map and score. \emph{Right:} the purifier $h_\psi$ is trained on the $k$ support images $x^{(\text{clean})}$
    paired with two types of perturbed counterparts -- Gaussian-noise
    augmentations $x^{(\text{noise})}$ and DP-generated
    adversarial samples $x^{(\text{adv})}$ -- by minimizing the two-term
    purifier loss $\mathcal{L}_\psi$.}
    \label{fig:defense_framework}
\end{figure}

\subsection{Defense Framework: PatchShift and FeaturePurifier}
\label{sec:defense}

We propose a two-stage backbone-training-free defense operating at two
complementary levels of the scoring pipeline: the input pixel space and
the feature space. Both components rely only on the $k$ few-shot
support images and on adversarial augmentations of those images. Figure~\ref{fig:defense_framework} provides an overview of the
robust pipeline and the FeaturePurifier training procedure.

\subsubsection{PatchShift (PS): Input-Level Misalignment with the Patch Grid}
\label{sec:patchshift}
DINOv2 partitions its input into a fixed grid of non-overlapping
$P \times P$ patches whose locations are determined by the image
origin. Adversarial perturbations crafted against a fixed grid are
tightly coupled to that grid: small spatial translations of the input
shift the perturbation relative to the patch boundaries, breaking the
alignment between the optimized perturbation and the encoder's
tokenization. Gu~\textit{et al.}~\cite{gu2022vision}
observed that this misalignment substantially attenuates the effect of
adversarial \emph{patch} perturbations against ViT classifiers. We adapt
the same geometric idea to the $\ell_\infty$ pixel-wise setting and the
$k$-NN anomaly detector, where the score depends on per-patch tokens
rather than a class prediction.

Let $x$ be a test image. We sample $K$ shifts
$\{(\Delta_y^{(k)}, \Delta_x^{(k)})\}_{k=1}^{K}$ with $\Delta^{(1)} = (0,0)$
and $\Delta^{(k)} \sim \mathrm{Unif}\{0,1,\dots,\Delta_{\max}\}^2$ for
$k \geq 2$. For each shift, we form the shifted-and-cropped image
\begin{equation}
\label{eq:shift_op}
x^{(k)} \;=\; \mathrm{Crop}_{P}\!\bigl(\mathrm{Shift}_{\Delta^{(k)}}(x)\bigr),
\end{equation}
where $\mathrm{Shift}_{\Delta^{(k)}}$ removes the top $\Delta_y^{(k)}$
rows and left $\Delta_x^{(k)}$ columns, and $\mathrm{Crop}_P$ crops to
the largest sub-image whose height and width are multiples of $P$. We
score each shifted image through the (purified, see~\S\ref{sec:purifier})
backbone and $k$-NN search, obtaining a per-patch distance map
$a^{(k)} \in \mathbb{R}^{N^{(k)}}$. Because the shift may change
$N^{(k)}$, we bilinearly resize each $a^{(k)}$ to the
unshifted grid size, yielding aligned maps
$\bar{a}^{(k)} \in \mathbb{R}^N$. The PS-aggregated anomaly
map is the element-wise median across shifts,
\begin{equation}
\label{eq:patchshift_median}
a^{\mathrm{PS}}(x) \;=\; \mathrm{median}_{k=1,\dots,K}\!\bigl(\bar{a}^{(k)}\bigr),
\end{equation}
followed by a small $3 \times 3$ spatial median filter for local
smoothing. The robust image-level score is
$s^{\mathrm{PS}}(x) = q(a^{\mathrm{PS}}(x))$ via
Eq.~\eqref{eq:mean_top1p}.

The element-wise median is preferred over the mean because adversarial
perturbations under \emph{some} shifts may remain partially aligned with
the patch grid and yield outlier-large distances; the median is robust
to such residual alignment, while the mean would propagate it into the
aggregate.

\begin{algorithm}[t]
\caption{Robust anomaly scoring (PatchShift~$\circ$~FeaturePurifier)}
\label{alg:defended_scoring}
\begin{algorithmic}[1]
\Require Test image $x$, frozen backbone $f_\theta$, memory bank $\mathcal{M}$, purifier $h_\psi$, shift count $K$, max shift $\Delta_{\max}$.
\State $\Delta^{(1)} \gets (0,0)$; sample $\Delta^{(k)} \sim \mathrm{Unif}\{0,\dots,\Delta_{\max}\}^2$ for $k = 2,\dots,K$.
\For{$k = 1, \dots, K$}
  \State $x^{(k)} \gets \mathrm{Crop}_{P}(\mathrm{Shift}_{\Delta^{(k)}}(x))$
  \State $(p_1, \dots, p_{N^{(k)}}) \gets f_\theta(x^{(k)})$
  \State $q_j \gets h_\psi(p_j)$ for $j = 1, \dots, N^{(k)}$
  \State $a_j^{(k)} \gets \tfrac{1}{2}\lVert \tilde{q}_j - \tilde{p}_{\mathrm{NN}(j)} \rVert_2^2$ where $p_{\mathrm{NN}(j)} \in \mathcal{M}$ is the nearest neighbor.
  \State $\bar{a}^{(k)} \gets \mathrm{Resize}(a^{(k)}, N)$
\EndFor
\State $a^{\mathrm{PS}} \gets \mathrm{median}_k(\bar{a}^{(k)})$
\State $a^{\mathrm{PS}} \gets \mathrm{MedianFilter}_{3\times3}(a^{\mathrm{PS}})$
\State \Return $s^{\mathrm{PS}}(x) \gets \mathrm{mean}(H_{0.01}(\bm{a}^{\mathrm{PS}}))$, $a^{\mathrm{PS}}$
\end{algorithmic}
\end{algorithm}

\subsubsection{FeaturePurifier (FP): Feature-Level Manifold Projection}
\label{sec:purifier}
PS attenuates the adversarial signal at the input level, but
some perturbation generally survives into the feature space. The
FeaturePurifier (FP), adapted from Dong~\textit{et al.}~\cite{dong2022improving}
in the few-shot adversarially-robust classification setting, learns a
mapping from perturbed to clean feature representations on the few-shot
support set. The mapping is then applied to every feature extracted at
test time, projecting it toward the manifold of clean nominal features
before $k$-NN search. 

The purifier $h_\psi: \mathbb{R}^D \to \mathbb{R}^D$ is a residual
two-layer MLP with hidden dimension $h$,
\begin{equation}
\label{eq:purifier_arch}
h_\psi(p) \;=\; p \,+\, V_2\,\sigma\!\bigl(V_1 p + c_1\bigr) + c_2,
\end{equation}
where $\sigma$ is the ReLU activation. The output weights
$V_2 \in \mathbb{R}^{D \times h}$ and bias
$c_2 \in \mathbb{R}^D$ are initialized to zero, so that $h_\psi$
initializes to the identity. This guarantees that, prior to training,
the purifier has no effect on the clean detector.

Training pairs of (clean, perturbed) features are constructed entirely
from the $k$ support images. For each support image $x^{(i)}$ we
extract its clean masked tokens $\{p_j^{(i)}\}_j$ via $f_\theta$.
We then generate two complementary sources of perturbed counterparts:
(i)~adversarial examples produced by the DP-driven PGD attack
of Section~\ref{sec:attack} applied to $x^{(i)}$ with a zero
ground-truth mask (since support images are all nominal); and
(ii)~$L_{\mathrm{noise}}$ Gaussian-noise augmentations
$x_\ell^{(i)} = \mathrm{clip}_{[0,1]}\!\bigl(x^{(i)} + \eta_\ell\bigr)$
with $\eta_\ell \sim \mathcal{N}(0,\,\sigma_{\mathrm{noise}}^2\, I)$.
Both sources are passed through $f_\theta$ to obtain perturbed feature
sets $\{\tilde{p}_j^{(i)}\}_j$, each paired with the same clean
feature $p_j^{(i)}$. The combination encourages the purifier to
generalize beyond a single perturbation distribution.

Let $\mathcal{P}_{\mathrm{train}} = \{(p_n, \tilde{p}_n)\}_{n=1}^{M_{\mathrm{p}}}$
denote the resulting set of (clean, perturbed) feature pairs over all
support images and noise / adversarial draws. The purifier is trained
to minimize the two-term mean-squared error
\begin{equation}
\label{eq:purifier_loss}
\mathcal{L}_{\mathrm{purify}}(\psi) \;=\; \frac{1}{M_{\mathrm{p}}}\sum_{n=1}^{M_{\mathrm{p}}}\bigl\lVert h_\psi(\tilde{p}_n) - p_n \bigr\rVert_2^2 \;+\; \frac{1}{M_{\mathrm{p}}}\sum_{n=1}^{M_{\mathrm{p}}}\bigl\lVert h_\psi(p_n) - p_n \bigr\rVert_2^2.
\end{equation}
The first term encourages the purifier to map perturbed features back
to their clean counterparts; the second term explicitly constrains the
purifier to leave clean features unchanged, preventing degradation of
clean-image performance.

At test time, the purifier is inserted between feature extraction and
$k$-NN search. For every patch token $p_j = f_\theta(x)_j$
the scoring pipeline computes $h_\psi(p_j)$, $\ell_2$-normalizes it,
and queries the nearest neighbor search. When PS is also active, $h_\psi$ is
applied independently to the features of each shifted view before the
$k$-NN search and before the median aggregation of
Eq.~\eqref{eq:patchshift_median}. The full robust scoring procedure is
summarized in Algorithm~\ref{alg:defended_scoring}.

\section{Experiments}
\label{sec:experiments}

\subsection{Datasets, Backbone, and Evaluation}
\label{sec:datasets_backbone_eval}
 
We evaluate on three industrial anomaly detection
benchmarks: MVTec-AD~\cite{bergmann2019mvtec}, VisA~\cite{zou2022spot},
and MPDD~\cite{jezek2021deep}, spanning 15, 12, and 6 categories
respectively. The three differ in clean-baseline difficulty due to
varying scene complexity, object orientation, and background
heterogeneity. We employ each dataset's official train/test split and
draw the few-shot support from the train split's nominal images only following \cite{damm2025anomalydino}.
 
The frozen feature extractor is DINOv2 ViT-S/14~\cite{oquab2023dinov2}
following ~\cite{damm2025anomalydino}, producing
$D = 384$ dimensional tokens at patch size $P = 14$. Images are resized
to a smaller edge of 448 pixels and center-cropped to the largest
multiple of $P$, yielding $N = 1024$ patches per image. Shot counts of
$k \in \{1, 2, 4, 8\}$ cover the full few-shot regime. $1$-nearest-neighbor
search runs over $\ell_2$-normalized features.
 
We report the image-level area under the
receiver-operator curve (AUROC), pixel-level AUROC and Per-Region Overlap (PRO) under clean, adversarial, and robust conditions. All numbers
are averaged across $3$ random seeds and object-balanced across categories within each dataset following the standard FSAD convention~\cite{damm2025anomalydino, jeong2023winclip}. Per-category results for all reported conditions and shot counts, are provided in the supplementary material.
We compare against PatchGuard~\cite{nafez2025patchguard} and
COBRA~\cite{mirzaei2025adversarially}, the current state of the art robust anomaly detection methods. All experiments run on a single NVIDIA RTX 6000.
 
\subsection{Attack, Defense, and Probe Configuration}
\label{sec:attack_defense_config}
 Following the convention adopted in adversarial anomaly
detection~\cite{nafez2025patchguard, mirzaei2025adversarially}, our
primary threat model is $\ell_\infty$ PGD-20 at $\epsilon = 8/255$ with random start
inside the $\ell_\infty$ ball. The surrogate probe DP is a three-layer MLP with hidden dimension $h = D = 384$. Training uses $n_{\mathrm{tr}} = 50$ nominal images
drawn from the tail of the sorted train split (disjoint from the
few-shot support, drawn from the head), trained for $100$ epochs with
Adam at learning rate $10^{-3}$, and regression
target $\log(1 + d_{\mathrm{NN}})$ per Eq.~\eqref{eq:probe_loss}.
We verify probe calibration via the Spearman rank correlation between
predicted and true $k$-NN distances on a held-out batch; correlations
exceed $0.85$ across all categories.

PS uses $K = 8$ shifted views with maximum shift
$\Delta_{\max} = 7$ pixels per axis (half the patch size), followed by
a $3 \times 3$ spatial median filter on the aggregated map. The
FP $h_\psi$ is a residual MLP with hidden dimension
$h = D = 384$, output layer zero-initialized to start at the identity,
trained for $100$ epochs with Adam at learning rate $10^{-3}$.
Training pairs per support image consist of $L_{\mathrm{noise}} = 4$
Gaussian-noise augmentations
($\sigma_{\mathrm{noise}} = 0.03$ in pixel $[0,1]$ space) plus one
PGD-attacked version generated by the DP-driven PGD attack
with $10$ inner iterations at the test-time $\epsilon$, contributing
to the two-term mean-squared error of Eq.~\eqref{eq:purifier_loss}.

\begin{table}[t]
\centering
\caption{Image-level detection AUROC (\%) under clean, adversarial,
\textcolor{gray}{clean}/robust in adversarial conditions across
shot counts. Adversarial attack is PGD-20 at $\epsilon = 8/255$ on our
DistanceProbe surrogate. All values are mean $\pm$ standard deviation
across $3$ random seeds.}
\label{tab:main_detection}
\setlength{\tabcolsep}{4pt}
\small
\resizebox{\linewidth}{!}{%
\begin{tabular}{llcccc}
\toprule
\textbf{Dataset} & \textbf{Condition} & \textbf{1-shot} & \textbf{2-shot} & \textbf{4-shot} & \textbf{8-shot} \\
\midrule
\multirow{3}{*}{MVTec-AD}
 & Clean         & 96.5{\scriptsize\,$\pm$\,0.5} & 96.7{\scriptsize\,$\pm$\,1.0} & 97.5{\scriptsize\,$\pm$\,0.1} & 98.0{\scriptsize\,$\pm$\,0.1} \\
 & Adversarial   & 20.8{\scriptsize\,$\pm$\,2.3} & 23.9{\scriptsize\,$\pm$\,3.2} & 25.8{\scriptsize\,$\pm$\,3.1} & 23.5{\scriptsize\,$\pm$\,3.0} \\
 & \textbf{Robust}
  & \textcolor{gray}{96.4{\scriptsize\,$\pm$\,0.7}} / \textbf{65.1}{\scriptsize\,$\pm$\,0.5}
  & \textcolor{gray}{96.8{\scriptsize\,$\pm$\,1.0}} / \textbf{72.9}{\scriptsize\,$\pm$\,1.1}
  & \textcolor{gray}{97.7{\scriptsize\,$\pm$\,0.2}} / \textbf{76.2}{\scriptsize\,$\pm$\,1.9}
  & \textcolor{gray}{98.2{\scriptsize\,$\pm$\,0.2}} / \textbf{82.6}{\scriptsize\,$\pm$\,1.2} \\
\midrule
\multirow{3}{*}{VisA}
 & Clean         & 85.7{\scriptsize\,$\pm$\,1.8} & 88.3{\scriptsize\,$\pm$\,2.1} & 91.2{\scriptsize\,$\pm$\,0.9} & 92.5{\scriptsize\,$\pm$\,0.1} \\
 & Adversarial   & 21.2{\scriptsize\,$\pm$\,1.7} & 22.6{\scriptsize\,$\pm$\,1.3} & 24.9{\scriptsize\,$\pm$\,1.5} & 24.4{\scriptsize\,$\pm$\,0.7} \\
 & \textbf{Robust}
  & \textcolor{gray}{85.5{\scriptsize\,$\pm$\,1.1}} / \textbf{51.6}{\scriptsize\,$\pm$\,0.9}
  & \textcolor{gray}{87.9{\scriptsize\,$\pm$\,1.4}} / \textbf{57.4}{\scriptsize\,$\pm$\,1.2}
  & \textcolor{gray}{89.5{\scriptsize\,$\pm$\,1.0}} / \textbf{61.8}{\scriptsize\,$\pm$\,1.0}
  & \textcolor{gray}{90.5{\scriptsize\,$\pm$\,0.3}} / \textbf{62.0}{\scriptsize\,$\pm$\,2.0} \\
\midrule
\multirow{3}{*}{MPDD}
 & Clean         & 74.8{\scriptsize\,$\pm$\,1.5} & 77.8{\scriptsize\,$\pm$\,2.7} & 78.7{\scriptsize\,$\pm$\,2.2} & 81.9{\scriptsize\,$\pm$\,2.0} \\
 & Adversarial   & 30.1{\scriptsize\,$\pm$\,0.2} & 32.6{\scriptsize\,$\pm$\,5.8} & 32.1{\scriptsize\,$\pm$\,1.0} & 34.1{\scriptsize\,$\pm$\,2.1} \\
 & \textbf{Robust}
  & \textcolor{gray}{72.3{\scriptsize\,$\pm$\,2.4}} / \textbf{52.1}{\scriptsize\,$\pm$\,2.2}
  & \textcolor{gray}{74.3{\scriptsize\,$\pm$\,4.5}} / \textbf{52.5}{\scriptsize\,$\pm$\,5.9}
  & \textcolor{gray}{76.3{\scriptsize\,$\pm$\,3.9}} / \textbf{59.4}{\scriptsize\,$\pm$\,4.1}
  & \textcolor{gray}{78.8{\scriptsize\,$\pm$\,1.4}} / \textbf{60.8}{\scriptsize\,$\pm$\,4.4} \\
\bottomrule
\end{tabular}%
}
\end{table}

\begin{table}[h]
\centering
\caption{Pixel-level localization PRO (\%) under clean, adversarial,
\textcolor{gray}{clean}/robust in adversarial conditions across shot
counts. Adversarial attack is PGD-20 at $\epsilon = 8/255$ on our
DistanceProbe surrogate. All values are mean $\pm$ standard deviation
across $3$ random seeds.}
\label{tab:main_localization}
\setlength{\tabcolsep}{4pt}
\small
\resizebox{\linewidth}{!}{%
\begin{tabular}{llcccc}
\toprule
\textbf{Dataset} & \textbf{Condition} & \textbf{1-shot} & \textbf{2-shot} & \textbf{4-shot} & \textbf{8-shot} \\
\midrule
\multirow{3}{*}{MVTec-AD}
 & Clean         & 91.7{\scriptsize\,$\pm$\,0.1} & 92.0{\scriptsize\,$\pm$\,0.3} & 92.4{\scriptsize\,$\pm$\,0.2} & 92.7{\scriptsize\,$\pm$\,0.0} \\
 & Adversarial   & 7.7{\scriptsize\,$\pm$\,1.4}  & 9.2{\scriptsize\,$\pm$\,2.1}  & 11.3{\scriptsize\,$\pm$\,2.0} & 11.8{\scriptsize\,$\pm$\,1.6} \\
 & \textbf{Robust}
  & \textcolor{gray}{91.3{\scriptsize\,$\pm$\,0.1}} / \textbf{53.4}{\scriptsize\,$\pm$\,0.8}
  & \textcolor{gray}{91.3{\scriptsize\,$\pm$\,0.4}} / \textbf{59.4}{\scriptsize\,$\pm$\,0.5}
  & \textcolor{gray}{91.9{\scriptsize\,$\pm$\,0.2}} / \textbf{64.4}{\scriptsize\,$\pm$\,1.3}
  & \textcolor{gray}{92.5{\scriptsize\,$\pm$\,0.1}} / \textbf{67.9}{\scriptsize\,$\pm$\,0.7} \\
\midrule
\multirow{3}{*}{VisA}
 & Clean         & 90.7{\scriptsize\,$\pm$\,0.6} & 91.7{\scriptsize\,$\pm$\,0.6} & 92.5{\scriptsize\,$\pm$\,0.2} & 93.2{\scriptsize\,$\pm$\,0.2} \\
 & Adversarial   & 17.2{\scriptsize\,$\pm$\,0.4} & 19.9{\scriptsize\,$\pm$\,3.7} & 20.7{\scriptsize\,$\pm$\,1.0} & 19.6{\scriptsize\,$\pm$\,0.8} \\
 & \textbf{Robust}
  & \textcolor{gray}{89.5{\scriptsize\,$\pm$\,0.7}} / \textbf{67.2}{\scriptsize\,$\pm$\,0.5}
  & \textcolor{gray}{90.4{\scriptsize\,$\pm$\,0.3}} / \textbf{68.8}{\scriptsize\,$\pm$\,0.7}
  & \textcolor{gray}{91.3{\scriptsize\,$\pm$\,0.1}} / \textbf{72.1}{\scriptsize\,$\pm$\,1.2}
  & \textcolor{gray}{92.1{\scriptsize\,$\pm$\,0.1}} / \textbf{73.3}{\scriptsize\,$\pm$\,0.4} \\
\midrule
\multirow{3}{*}{MPDD}
 & Clean         & 88.4{\scriptsize\,$\pm$\,0.6} & 89.7{\scriptsize\,$\pm$\,0.8} & 90.5{\scriptsize\,$\pm$\,0.9} & 91.0{\scriptsize\,$\pm$\,0.5} \\
 & Adversarial   & 4.0{\scriptsize\,$\pm$\,1.4}  & 6.0{\scriptsize\,$\pm$\,1.3}  & 5.9{\scriptsize\,$\pm$\,2.2}  & 6.5{\scriptsize\,$\pm$\,1.1} \\
 & \textbf{Robust}
  & \textcolor{gray}{87.1{\scriptsize\,$\pm$\,0.7}} / \textbf{48.3}{\scriptsize\,$\pm$\,2.9}
  & \textcolor{gray}{87.2{\scriptsize\,$\pm$\,1.9}} / \textbf{54.6}{\scriptsize\,$\pm$\,7.0}
  & \textcolor{gray}{84.4{\scriptsize\,$\pm$\,3.6}} / \textbf{59.2}{\scriptsize\,$\pm$\,6.4}
  & \textcolor{gray}{86.4{\scriptsize\,$\pm$\,2.9}} / \textbf{62.9}{\scriptsize\,$\pm$\,4.6} \\
\bottomrule
\end{tabular}%
}
\end{table}

\subsection{Robustness Across Few Shot Settings}
\label{sec:main_results}
Tables~\ref{tab:main_detection} and~\ref{tab:main_localization} report
image-level detection AUROC and pixel-level localization PRO under
clean, adversarial, clean-defended, and robust (adversarial-defended)
conditions. We report PRO rather than
pixel-level AUROC because nominal/anomalous pixel imbalance reaches
$97$--$99\%$ across these
datasets~\cite{damm2025anomalydino, li2024musc, rafiei2023pixel}; pixel-level AUROC
is reported in the supplementary material. The DP-driven PGD
attack produces strong bidirectional ranking inversion, dropping
detection AUROC from $\geq 75\%$ clean to $\approx 21$--$34\%$ and
localization PRO from $\approx 90\%$ to under $18\%$ on every dataset,
confirming that the regression-trained probe is a faithful surrogate
for the non-parametric $k$-NN scorer. The combined defense recovers substantial robustness across all datasets and shot counts: at $k = 4$, detection AUROC reaches
$76.2/61.8/59.4\%$ on MVTec-AD/VisA/MPDD, and localization PRO reaches
$64.4/72.1/59.2\%$. Recovery improves monotonically with shot count
in nearly every cell, indicating the FP benefits from a
richer support distribution. Clean performance is preserved: defended
clean AUROC matches undefended clean within $0.2$, $2.0$, and $3.1$
percentage points on the three datasets respectively, with the largest
single drop occurring on MPDD localization PRO at $k=4$ ($90.5\%$ to
$84.4\%$). The absolute defended numbers vary across datasets but the
qualitative pattern is consistent throughout. See supplementary for per-category breakdowns.

\subsection{Comparison with state-of-the-art Robust Anomaly Detection Methods}
\label{sec:comparison}

Table~\ref{tab:comparison} compares our method against
COBRA~\cite{mirzaei2025adversarially} and
PatchGuard~\cite{nafez2025patchguard} on MVTec-AD, VisA, and MPDD.
Our method operates at $k = 4$ shots with a frozen DINOv2 ViT-S/14
backbone; both baselines use the full training set with their original
backbones (ResNet-18 for COBRA, a from-scratch ViT for PatchGuard).
Regarding backbone disparity, prior methods invest
training-side compute in the feature extractor, while we leverage
DINOv2's strong pretrained representations and invest only in a
lightweight purifier MLP on $k$ support images. All baseline
numbers were re-run under the PGD-20 attack configuration at the
$\epsilon$ each method recommended and was originally evaluated against: $2/255$ for
COBRA and $8/255$ for PatchGuard and our method, a $4\times$ stronger
budget. For this comparison we report pixel-level AUROC for
localization rather than PRO (as in our main results in
Table~\ref{tab:main_localization}) to match the metric used in
the baselines' original publications and ensure fair comparison. On image-level detection at $\epsilon=8/255$, our method
exceeds PatchGuard on MVTec-AD ($97.7/76.2$ vs $92.7/68.3$ clean/robust
AUROC), while PatchGuard achieves stronger robust AUROC on VisA and
MPDD where DINOv2's frozen features underperform PatchGuard's trained
backbone even on clean inputs (notably MPDD, where our clean AUROC trails PatchGuard's by $9.5$ points). On pixel-level localization our method outperforms
PatchGuard on both MVTec-AD ($81.0$ vs $75.5$ robust) and VisA ($91.2$
vs $85.1$ robust), and ties closely on MPDD clean localization
($93.9$ vs $94.0$). Pixel-level localization is the more practically
informative metric for industrial inspection where the goal is to
identify defective regions rather than merely flag images, and our
defense's recovery is strongest in this regime. Against COBRA on
MVTec-AD, our method achieves comparable robust AUROC ($76.2$ vs
$75.1$) under a $4\times$ stronger attack while substantially
exceeding COBRA's clean performance ($97.7$ vs $89.1$), indicating
that the gain comes from a stronger underlying detector rather than
from a weaker attack.

\begin{table}[t]
\centering
\caption{Comparison against published adversarially robust anomaly
detection methods. Cells show \textcolor{gray}{Clean}\,/\,Robust AUROC
under PGD-20. COBRA was originally evaluated at $\epsilon = 2/255$;
PatchGuard and our method use the stronger $\epsilon = 8/255$ budget.}
\label{tab:comparison}
\setlength{\tabcolsep}{3pt}
\small
\begin{tabular}{llcccccc}
\toprule
\textbf{Dataset} & \textbf{Method} & \textbf{Venue} & \textbf{Backbone} & \textbf{Shots} & \textbf{\makecell{Image-level\\AUROC (\%)}} & \textbf{\makecell{Pixel-level\\AUROC (\%)}} \\
\midrule
\multirow{3}{*}{MVTec-AD}
 & COBRA~\cite{mirzaei2025adversarially}     & ICLR'25 & ResNet-18 & full & \textcolor{gray}{89.1}\,/\,75.1 & --- \\
 & PatchGuard~\cite{nafez2025patchguard}     & CVPR'25 & ViT       & full & \textcolor{gray}{92.7}\,/\,68.3 & \textcolor{gray}{96.5}\,/\,75.5 \\
 & \textbf{PS+FP}                     & Ours & DINOv2    & 4    & \textcolor{gray}{97.7}\,/\,\textbf{76.2} & \textcolor{gray}{97.3}\,/\,\textbf{81.0} \\
\midrule
\multirow{3}{*}{VisA}
 & COBRA~\cite{mirzaei2025adversarially}     & ICLR'25 & ResNet-18 & full & \textcolor{gray}{75.2}\,/\,73.8 & --- \\
 & PatchGuard~\cite{nafez2025patchguard}     & CVPR'25 & ViT       & full & \textcolor{gray}{90.8}\,/\,\textbf{75.8} & \textcolor{gray}{97.2}\,/\,85.1 \\
 & \textbf{PS+FP}                     & Ours & DINOv2    & 4    & \textcolor{gray}{89.5}\,/\,61.8 & \textcolor{gray}{98.3}\,/\,\textbf{91.2} \\
\midrule
\multirow{2}{*}{MPDD}
 & PatchGuard~\cite{nafez2025patchguard}     & CVPR'25 & ViT       & full & \textcolor{gray}{85.8}\,/\,\textbf{70.6} & \textcolor{gray}{94.0}\,/\,\textbf{83.7} \\
 & \textbf{PS+FP}                     & Ours & DINOv2    & 4    & \textcolor{gray}{76.3}\,/\,59.4 & \textcolor{gray}{93.9}\,/\,77.2 \\
\bottomrule
\end{tabular}
\end{table}

\subsection{Adaptive Attack Evaluation}
\label{sec:adaptive_eval}

To verify that our method's robustness is not an artifact of gradient
obfuscation~\cite{athalye2018obfuscated}, we evaluate against a fully
adaptive attacker that differentiates through every component of the
robust pipeline. The adaptive attack combines Expectation over
Transformation (EOT) to handle
PS's test-time random sampling, with Backward Pass
Differentiable Approximation (BPDA) adapted from \cite{tramer2020adaptive}, to
provide stable gradients through the discrete shift operation with the
FP being differentiable and included directly in the attack
graph. The per-PGD-step gradient is averaged over $n_{\mathrm{EOT}} =
10$ shift samples with shift magnitudes matching the defender's
$\Delta_{\max} = 7$. We also compare against
single-step FGSM \cite{goodfellow2014explaining} (a one-step gradient sign attack) and
$\ell_\infty$-PGD-20, all at $\epsilon = 8/255$. Table~\ref{tab:adaptive} reports robust AUROC under all three attacks. EOT-BPDA-PGD is consistently the strongest attack on MVTec-AD and MPDD, producing robust image-level AUROC of $67.1/62.5/55.4$\% on MVTec-AD/VisA/MPDD, $5$--$10$ points below PGD-20 on most
cells except on VisA where it is slightly weaker. This gap reflects the adaptive attacker's strictly better
information rather than a defense breakdown as the defense continues to
recover far above the attack floor. FGSM produces the weakest
attack across all settings, confirming that the attack benefits from
iterative gradient information rather than gradient masking. We note that AutoAttack~\cite{croce2020reliable} is
not directly applicable in our setting as it assumes a parametric
classification head whose logits can be queried and differentiated. Adapting AutoAttack to our setting would require the same DP surrogate, in which case the comparison would reduce to PGD-20 against EOT-BPDA-PGD as we report here.

\begin{table}[t]
\centering
\caption{Robust AUROC and PRO (\%) under three attacks at $\epsilon = 8/255$:
single-step FGSM, $\ell_\infty$-PGD-20, and EOT-BPDA-PGD \cite{tramer2020adaptive}.}
\label{tab:adaptive}
\setlength{\tabcolsep}{4pt}
\small
\begin{tabular}{llcccc}
\toprule
\textbf{Dataset} & \textbf{Metric} & \textbf{Clean} & \textbf{FGSM} & \textbf{EOT-BPDA-PGD} & $\boldsymbol{\ell_\infty}$\textbf{-PGD} \\
\midrule
\multirow{2}{*}{MVTec-AD}
 & Image AUROC & 97.7 & 78.2 & \textbf{67.1} & 76.2 \\
 & PRO         & 91.9 & 72.5 & \textbf{64.9} & 64.4 \\
\midrule
\multirow{2}{*}{VisA}
 & Image AUROC & 89.5 & 73.3 & 62.5 & \textbf{61.8} \\
 & PRO         & 91.3 & 81.7 & 79.1 & \textbf{72.1} \\
\midrule
\multirow{2}{*}{MPDD}
 & Image AUROC & 76.3 & 61.6 & \textbf{55.4} & 59.4 \\
 & PRO         & 84.4 & 57.3 & \textbf{52.1} & 59.2 \\
\bottomrule
\end{tabular}
\end{table}

\subsection{Ablation Study}
\label{sec:ablation}

\textbf{Defense Component Ablation.} Table~\ref{tab:ablation_components} reports robust AUROC under PGD-20
attack with each defense component active independently and together.
The two components contribute differently: FeaturePurifier alone yields
the strongest image-level detection ($78.4/62.5$\% on MVTec-AD/VisA),
while PatchShift alone provides only modest detection recovery but
preserves spatial structure in the anomaly map. Their combination
sacrifices a small amount of image-level detection ($2.2$ points on
MVTec-AD) but substantially improves pixel-level localization
($+11.6$ PRO on MVTec-AD, $+12.5$ on VisA), reflecting the
complementary mechanisms: feature-space purification recovers per-patch
score reliability that drives image level detection, while input-space
shift aggregation smooths the spatial anomaly map that drives
localization. The combined defense is therefore preferable for
industrial inspection settings, where localizing defective regions
is the primary objective.

\begin{table}[t]
\centering
\caption{Defense component ablation: robust AUROC and PRO (\%) under PGD-20
attack at $\epsilon=8/255$, $k=4$ shots.}
\label{tab:ablation_components}
\setlength{\tabcolsep}{4pt}
\small
\begin{tabular}{lcccc}
\toprule
& \multicolumn{2}{c}{\textbf{MVTec-AD}} & \multicolumn{2}{c}{\textbf{VisA}} \\
\cmidrule(lr){2-3} \cmidrule(lr){4-5}
\textbf{Method} & \textbf{Image AUROC} & \textbf{PRO} & \textbf{Image AUROC} & \textbf{PRO} \\
\midrule
PS only     & 53.7 & 40.1 & 46.5 & 62.0 \\
FP only     & \textbf{78.4} & 52.8 & \textbf{62.5} & 59.6 \\
PS+FP (Ours) & 76.2 & \textbf{64.4} & 61.8 & \textbf{72.1} \\
\bottomrule
\end{tabular}
\end{table}

\begin{figure}[t]
\centering
\includegraphics[width=0.8\linewidth]{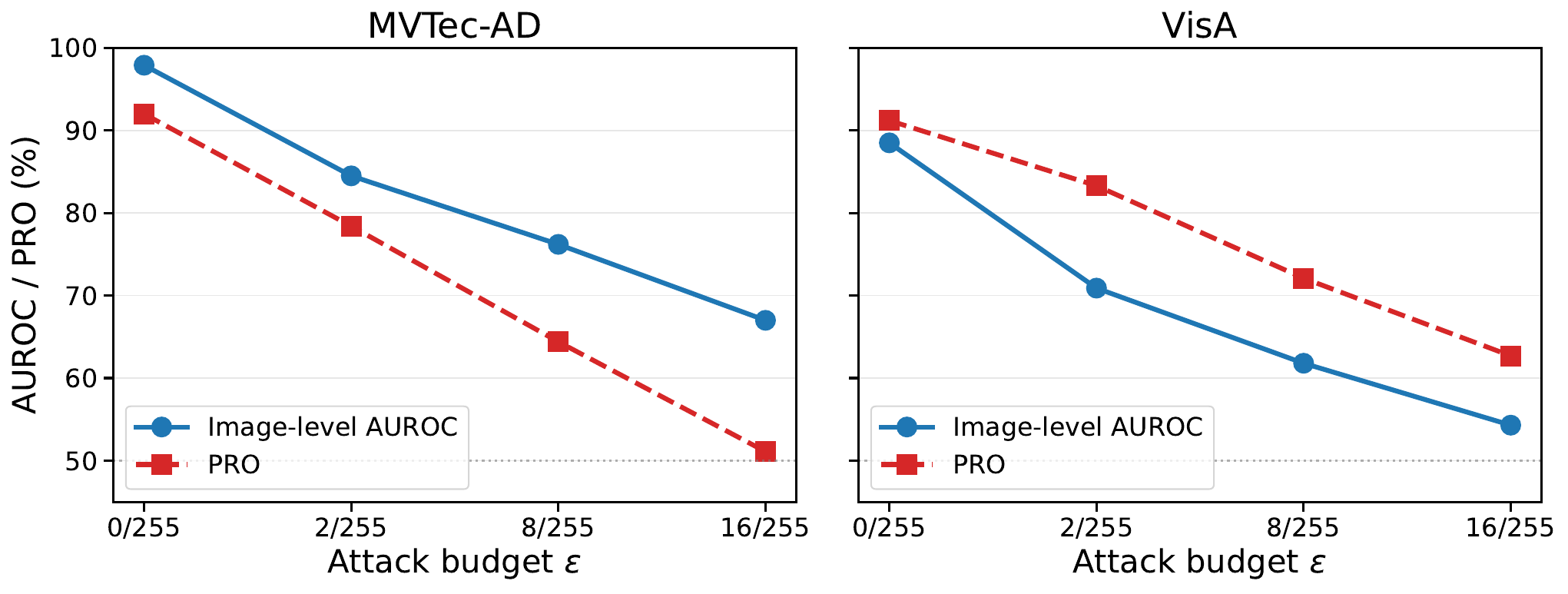}
\caption{Robust AUROC and PRO under PGD-20 attack as the attack
budget $\epsilon$ varies, at $k=4$ shots. Both metrics degrade
gracefully on both datasets with no abrupt collapse.}
\label{fig:eps_sweep}
\end{figure}

\noindent \textbf{Robustness Across Attack Budgets.} Figure~\ref{fig:eps_sweep} reports robust image-level AUROC and
pixel-level PRO as a function of the $\ell_\infty$ attack budget
$\epsilon \in \{0, 2, 8, 16\}/255$ at $k = 4$ shots. Both metrics degrade
gracefully with $\epsilon$ on both datasets, with no abrupt collapse:
image-level AUROC drops from $97.9/88.5$\% at $\epsilon=0$ to
$67.0/54.3$\% at $\epsilon=16/255$ on MVTec-AD and VisA respectively,
a $2\times$ stronger attack than the primary $\epsilon=8/255$ setting
used in our headline results. The smooth, monotone degradation
indicates that the defense provides progressive rather than fragile
robustness; the absence of a sharp inflection point at any specific
$\epsilon$ is consistent with genuine robustness rather than gradient
masking, since gradient-obfuscated defenses typically exhibit
discontinuous transitions where the attack suddenly succeeds beyond a
threshold~\cite{athalye2018obfuscated}. Notably, even at
$\epsilon=16/255$ -- double the budget at which the defense was
configured -- robust AUROC and PRO remain well above the inversion
floor on both datasets, indicating the defense does not overfit to a
specific attack budget.
\section{Conclusion and Future Work}
\label{sec:conclusion}
We presented the first adversarial robustness framework for few-shot
anomaly detection with frozen vision foundation models. The
\emph{DistanceProbe} surrogate enables strong PGD attacks on
non-parametric $k$-NN scorers, and our two-stage \emph{PatchShift +
FeaturePurifier} defense recovers substantial robustness on MVTec-AD,
VisA, and MPDD while preserving clean accuracy and matching full-shot
adversarially-trained baselines on pixel-level localization at
$k = 4$. The defense holds under adaptive EOT-BPDA attack, ruling out
gradient obfuscation.

\noindent \textbf{Limitations.} Defense recovery is weaker on MPDD's image-level
detection than on MVTec-AD or VisA, primarily reflecting the clean
baseline disparity where DINOv2's frozen features underperform
PatchGuard's trained backbone before any attack rather than a
defense-side weakness; the adaptive EOT-BPDA attacker also produces
$5$--$10$ AUROC point degradation over vanilla PGD-20, indicating
meaningful but not absolute robustness.

\noindent \textbf{Future Work.} Future directions include extending the
framework to other multi-modal pretrained backbones
(CLIP~\cite{radford2021learning}, MAE~\cite{bachmann2022multimae}), enriching the
purifier's training distribution with adaptive-attack samples, and
addressing physical-world perturbations beyond the $\ell_\infty$
threat model studied here.

\bibliography{egbib}

@inproceedings{roth2022towards,
  title={Towards total recall in industrial anomaly detection},
  author={Roth, Karsten and Pemula, Latha and Zepeda, Joaquin and Sch{\"o}lkopf, Bernhard and Brox, Thomas and Gehler, Peter},
  booktitle={Proceedings of the IEEE/CVF conference on computer vision and pattern recognition},
  pages={14318--14328},
  year={2022}
}

@inproceedings{damm2025anomalydino,
  title={Anomalydino: Boosting patch-based few-shot anomaly detection with dinov2},
  author={Damm, Simon and Laszkiewicz, Mike and Lederer, Johannes and Fischer, Asja},
  booktitle={2025 IEEE/CVF Winter Conference on Applications of Computer Vision (WACV)},
  pages={1319--1329},
  year={2025},
  organization={IEEE}
}

@article{oquab2023dinov2,
  title={Dinov2: Learning robust visual features without supervision},
  author={Oquab, Maxime and Darcet, Timoth{\'e}e and Moutakanni, Th{\'e}o and Vo, Huy and Szafraniec, Marc and Khalidov, Vasil and Fernandez, Pierre and Haziza, Daniel and Massa, Francisco and El-Nouby, Alaaeldin and others},
  journal={arXiv preprint arXiv:2304.07193},
  year={2023}
}

@inproceedings{nafez2025patchguard,
  title={Patchguard: Adversarially robust anomaly detection and localization through vision transformers and pseudo anomalies},
  author={Nafez, Mojtaba and Koochakian, Amirhossein and Maleki, Arad and Habibi, Jafar and Rohban, Mohammad Hossein},
  booktitle={Proceedings of the Computer Vision and Pattern Recognition Conference},
  pages={20383--20394},
  year={2025}
}

@article{madry2017towards,
  title={Towards deep learning models resistant to adversarial attacks},
  author={Madry, Aleksander and Makelov, Aleksandar and Schmidt, Ludwig and Tsipras, Dimitris and Vladu, Adrian},
  journal={arXiv preprint arXiv:1706.06083},
  year={2017}
}

@article{xu2020adversarial,
  title={Adversarial attacks and defenses in images, graphs and text: A review},
  author={Xu, Han and Ma, Yao and Liu, Hao-Chen and Deb, Debayan and Liu, Hui and Tang, Ji-Liang and Jain, Anil K},
  journal={International journal of automation and computing},
  volume={17},
  number={2},
  pages={151--178},
  year={2020},
  publisher={Springer}
}

@article{yuan2019adversarial,
  title={Adversarial examples: Attacks and defenses for deep learning},
  author={Yuan, Xiaoyong and He, Pan and Zhu, Qile and Li, Xiaolin},
  journal={IEEE transactions on neural networks and learning systems},
  volume={30},
  number={9},
  pages={2805--2824},
  year={2019},
  publisher={IEEE}
}

@inproceedings{
mirzaei2025adversarially,
title={Adversarially Robust Anomaly Detection through Spurious Negative Pair Mitigation},
author={Hossein Mirzaei and Mojtaba Nafez and Jafar Habibi and Mohammad Sabokrou and Mohammad Hossein Rohban},
booktitle={The Thirteenth International Conference on Learning Representations},
year={2025},
url={https://openreview.net/forum?id=t8fu5m8R5m}
}

@inproceedings{gu2022vision,
  title={Are vision transformers robust to patch perturbations?},
  author={Gu, Jindong and Tresp, Volker and Qin, Yao},
  booktitle={European Conference on Computer Vision},
  pages={404--421},
  year={2022},
  organization={Springer}
}

@inproceedings{dong2022improving,
  title={Improving adversarially robust few-shot image classification with generalizable representations},
  author={Dong, Junhao and Wang, Yuan and Lai, Jian-Huang and Xie, Xiaohua},
  booktitle={Proceedings of the IEEE/CVF Conference on Computer Vision and Pattern Recognition},
  pages={9025--9034},
  year={2022}
}

@inproceedings{bergmann2019mvtec,
  title={MVTec AD--A comprehensive real-world dataset for unsupervised anomaly detection},
  author={Bergmann, Paul and Fauser, Michael and Sattlegger, David and Steger, Carsten},
  booktitle={Proceedings of the IEEE/CVF conference on computer vision and pattern recognition},
  pages={9592--9600},
  year={2019}
}

@inproceedings{zou2022spot,
  title={Spot-the-difference self-supervised pre-training for anomaly detection and segmentation},
  author={Zou, Yang and Jeong, Jongheon and Pemula, Latha and Zhang, Dongqing and Dabeer, Onkar},
  booktitle={European conference on computer vision},
  pages={392--408},
  year={2022},
  organization={Springer}
}

@inproceedings{jezek2021deep,
  title={Deep learning-based defect detection of metal parts: evaluating current methods in complex conditions},
  author={Jezek, Stepan and Jonak, Martin and Burget, Radim and Dvorak, Pavel and Skotak, Milos},
  booktitle={2021 13th International congress on ultra modern telecommunications and control systems and workshops (ICUMT)},
  pages={66--71},
  year={2021},
  organization={IEEE}
}

@article{cao2024adversarially,
  title={Adversarially robust industrial anomaly detection through diffusion model},
  author={Cao, Yuanpu and Lin, Lu and Chen, Jinghui},
  journal={arXiv preprint arXiv:2408.04839},
  year={2024}
}

@inproceedings{mirzaei2025lyapunov,
  title={Adversarially robust out-of-distribution detection using lyapunov-stabilized embeddings},
  author={Mirzaei Sadeghlou, Hossein and Mathis, Mackenzie},
  booktitle={International Conference on Learning Representations},
  volume={2025},
  pages={21056--21073},
  year={2025}
}

@inproceedings{jeong2023winclip,
  title={Winclip: Zero-/few-shot anomaly classification and segmentation},
  author={Jeong, Jongheon and Zou, Yang and Kim, Taewan and Zhang, Dongqing and Ravichandran, Avinash and Dabeer, Onkar},
  booktitle={Proceedings of the IEEE/CVF conference on computer vision and pattern recognition},
  pages={19606--19616},
  year={2023}
}

@inproceedings{li2024musc,
  title={Musc: Zero-shot industrial anomaly classification and segmentation with mutual scoring of the unlabeled images},
  author={Li, Xurui and Huang, Ziming and Xue, Feng and Zhou, Yu},
  booktitle={The Twelfth International Conference on Learning Representations},
  year={2024}
}

@article{rafiei2023pixel,
  title={On pixel-level performance assessment in anomaly detection},
  author={Rafiei, Mehdi and Breckon, Toby P and Iosifidis, Alexandros},
  journal={arXiv preprint arXiv:2310.16435},
  year={2023}
}

@inproceedings{athalye2018obfuscated,
  title={Obfuscated gradients give a false sense of security: Circumventing defenses to adversarial examples},
  author={Athalye, Anish and Carlini, Nicholas and Wagner, David},
  booktitle={International conference on machine learning},
  pages={274--283},
  year={2018},
  organization={PMLR}
}

@article{tramer2020adaptive,
  title={On adaptive attacks to adversarial example defenses},
  author={Tramer, Florian and Carlini, Nicholas and Brendel, Wieland and Madry, Aleksander},
  journal={Advances in neural information processing systems},
  volume={33},
  pages={1633--1645},
  year={2020}
}

@article{goodfellow2014explaining,
  title={Explaining and harnessing adversarial examples},
  author={Goodfellow, Ian J and Shlens, Jonathon and Szegedy, Christian},
  journal={arXiv preprint arXiv:1412.6572},
  year={2014}
}

@inproceedings{siddiqui2019detecting,
  title={Detecting cyber attacks using anomaly detection with explanations and expert feedback},
  author={Siddiqui, Md Amran and Stokes, Jack W and Seifert, Christian and Argyle, Evan and McCann, Robert and Neil, Joshua and Carroll, Justin},
  booktitle={ICASSP 2019-2019 IEEE International Conference on Acoustics, Speech and Signal Processing (ICASSP)},
  pages={2872--2876},
  year={2019},
  organization={IEEE}
}

@article{fernando2021deep,
  title={Deep learning for medical anomaly detection--a survey},
  author={Fernando, Tharindu and Gammulle, Harshala and Denman, Simon and Sridharan, Sridha and Fookes, Clinton},
  journal={ACM computing surveys (CSUR)},
  volume={54},
  number={7},
  pages={1--37},
  year={2021},
  publisher={ACM New York, NY, USA}
}

@article{tian2023self,
  title={Self-supervised pseudo multi-class pre-training for unsupervised anomaly detection and segmentation in medical images},
  author={Tian, Yu and Liu, Fengbei and Pang, Guansong and Chen, Yuanhong and Liu, Yuyuan and Verjans, Johan W and Singh, Rajvinder and Carneiro, Gustavo},
  journal={Medical image analysis},
  volume={90},
  pages={102930},
  year={2023},
  publisher={Elsevier}
}

@article{liu2024deep,
  title={Deep industrial image anomaly detection: A survey},
  author={Liu, Jiaqi and Xie, Guoyang and Wang, Jinbao and Li, Shangnian and Wang, Chengjie and Zheng, Feng and Jin, Yaochu},
  journal={Machine Intelligence Research},
  volume={21},
  number={1},
  pages={104--135},
  year={2024},
  publisher={Springer}
}

@article{bergmann2022beyond,
  title={Beyond dents and scratches: Logical constraints in unsupervised anomaly detection and localization},
  author={Bergmann, Paul and Batzner, Kilian and Fauser, Michael and Sattlegger, David and Steger, Carsten},
  journal={International Journal of Computer Vision},
  volume={130},
  number={4},
  pages={947--969},
  year={2022},
  publisher={Springer}
}

@inproceedings{li2021cutpaste,
  title={Cutpaste: Self-supervised learning for anomaly detection and localization},
  author={Li, Chun-Liang and Sohn, Kihyuk and Yoon, Jinsung and Pfister, Tomas},
  booktitle={Proceedings of the IEEE/CVF conference on computer vision and pattern recognition},
  pages={9664--9674},
  year={2021}
}

@inproceedings{defard2021padim,
  title={Padim: a patch distribution modeling framework for anomaly detection and localization},
  author={Defard, Thomas and Setkov, Aleksandr and Loesch, Angelique and Audigier, Romaric},
  booktitle={International conference on pattern recognition},
  pages={475--489},
  year={2021},
  organization={Springer}
}

@inproceedings{mousakhan2024anomaly,
  title={Anomaly detection with conditioned denoising diffusion models},
  author={Mousakhan, Arian and Brox, Thomas and Tayyub, Jawad},
  booktitle={DAGM German Conference on Pattern Recognition},
  pages={181--195},
  year={2024},
  organization={Springer}
}

@inproceedings{batzner2024efficientad,
  title={Efficientad: Accurate visual anomaly detection at millisecond-level latencies},
  author={Batzner, Kilian and Heckler, Lars and K{\"o}nig, Rebecca},
  booktitle={Proceedings of the IEEE/CVF winter conference on applications of computer vision},
  pages={128--138},
  year={2024}
}

@inproceedings{hyun2024reconpatch,
  title={Reconpatch: Contrastive patch representation learning for industrial anomaly detection},
  author={Hyun, Jeeho and Kim, Sangyun and Jeon, Giyoung and Kim, Seung Hwan and Bae, Kyunghoon and Kang, Byung Jun},
  booktitle={Proceedings of the IEEE/CVF winter conference on applications of computer vision},
  pages={2052--2061},
  year={2024}
}

@inproceedings{fang2023fastrecon,
  title={Fastrecon: Few-shot industrial anomaly detection via fast feature reconstruction},
  author={Fang, Zheng and Wang, Xiaoyang and Li, Haocheng and Liu, Jiejie and Hu, Qiugui and Xiao, Jimin},
  booktitle={Proceedings of the IEEE/CVF International Conference on Computer Vision},
  pages={17481--17490},
  year={2023}
}

@inproceedings{zhu2024toward,
  title={Toward generalist anomaly detection via in-context residual learning with few-shot sample prompts},
  author={Zhu, Jiawen and Pang, Guansong},
  booktitle={Proceedings of the IEEE/CVF conference on computer vision and pattern recognition},
  pages={17826--17836},
  year={2024}
}

@inproceedings{gu2025univad,
  title={Univad: A training-free unified model for few-shot visual anomaly detection},
  author={Gu, Zhaopeng and Zhu, Bingke and Zhu, Guibo and Chen, Yingying and Tang, Ming and Wang, Jinqiao},
  booktitle={Proceedings of the Computer Vision and Pattern Recognition Conference},
  pages={15194--15203},
  year={2025}
}

@inproceedings{huang2022registration,
  title={Registration based few-shot anomaly detection},
  author={Huang, Chaoqin and Guan, Haoyan and Jiang, Aofan and Zhang, Ya and Spratling, Michael and Wang, Yan-Feng},
  booktitle={European conference on computer vision},
  pages={303--319},
  year={2022},
  organization={Springer}
}

@inproceedings{zhou2024anomalyclip,
  title={Anomalyclip: Object-agnostic prompt learning for zero-shot anomaly detection},
  author={Zhou, Qihang and Pang, Guansong and Tian, Yu and He, Shibo and Chen, Jiming},
  booktitle={International Conference on Learning Representations},
  volume={2024},
  pages={49705--49737},
  year={2024}
}

@inproceedings{cao2024adaclip,
  title={Adaclip: Adapting clip with hybrid learnable prompts for zero-shot anomaly detection},
  author={Cao, Yunkang and Zhang, Jiangning and Frittoli, Luca and Cheng, Yuqi and Shen, Weiming and Boracchi, Giacomo},
  booktitle={European conference on computer vision},
  pages={55--72},
  year={2024},
  organization={Springer}
}

@article{xie2023pushing,
  title={Pushing the limits of fewshot anomaly detection in industry vision: Graphcore},
  author={Xie, Guoyang and Wang, Jinbao and Liu, Jiaqi and Zheng, Feng and Jin, Yaochu},
  journal={arXiv preprint arXiv:2301.12082},
  year={2023}
}

@inproceedings{goodge2021robustness,
  title={Robustness of autoencoders for anomaly detection under adversarial impact},
  author={Goodge, Adam and Hooi, Bryan and Ng, See Kiong and Ng, Wee Siong},
  booktitle={Proceedings of the twenty-ninth international conference on international joint conferences on artificial intelligence},
  pages={1244--1250},
  year={2021}
}

@article{akhtar2018threat,
  title={Threat of adversarial attacks on deep learning in computer vision: A survey},
  author={Akhtar, Naveed and Mian, Ajmal},
  journal={Ieee Access},
  volume={6},
  pages={14410--14430},
  year={2018},
  publisher={IEEE}
}

@article{ruff2021unifying,
  title={A unifying review of deep and shallow anomaly detection},
  author={Ruff, Lukas and Kauffmann, Jacob R and Vandermeulen, Robert A and Montavon, Gr{\'e}goire and Samek, Wojciech and Kloft, Marius and Dietterich, Thomas G and M{\"u}ller, Klaus-Robert},
  journal={Proceedings of the IEEE},
  volume={109},
  number={5},
  pages={756--795},
  year={2021},
  publisher={IEEE}
}

@article{yang2024generalized,
  title={Generalized out-of-distribution detection: A survey},
  author={Yang, Jingkang and Zhou, Kaiyang and Li, Yixuan and Liu, Ziwei},
  journal={International Journal of Computer Vision},
  volume={132},
  number={12},
  pages={5635--5662},
  year={2024},
  publisher={Springer}
}

@incollection{eskin2002geometric,
  title={A geometric framework for unsupervised anomaly detection: Detecting intrusions in unlabeled data},
  author={Eskin, Eleazar and Arnold, Andrew and Prerau, Michael and Portnoy, Leonid and Stolfo, Sal},
  booktitle={Applications of data mining in computer security},
  pages={77--101},
  year={2002},
  publisher={Springer}
}

@article{bergman2020deep,
  title={Deep nearest neighbor anomaly detection},
  author={Bergman, Liron and Cohen, Niv and Hoshen, Yedid},
  journal={arXiv preprint arXiv:2002.10445},
  year={2020}
}

@inproceedings{yi2020patch,
  title={Patch svdd: Patch-level svdd for anomaly detection and segmentation},
  author={Yi, Jihun and Yoon, Sungroh},
  booktitle={Proceedings of the Asian conference on computer vision},
  year={2020}
}

@article{cohen2020sub,
  title={Sub-image anomaly detection with deep pyramid correspondences},
  author={Cohen, Niv and Hoshen, Yedid},
  journal={arXiv preprint arXiv:2005.02357},
  year={2020}
}

@inproceedings{he2016deep,
  title={Deep residual learning for image recognition},
  author={He, Kaiming and Zhang, Xiangyu and Ren, Shaoqing and Sun, Jian},
  booktitle={Proceedings of the IEEE conference on computer vision and pattern recognition},
  pages={770--778},
  year={2016}
}

@article{zagoruyko2016wide,
  title={Wide residual networks},
  author={Zagoruyko, Sergey and Komodakis, Nikos},
  journal={arXiv preprint arXiv:1605.07146},
  year={2016}
}

@article{dosovitskiy2020image,
  title={An image is worth 16x16 words: Transformers for image recognition at scale},
  author={Dosovitskiy, Alexey and Beyer, Lucas and Kolesnikov, Alexander and Weissenborn, Dirk and Zhai, Xiaohua and Unterthiner, Thomas and Dehghani, Mostafa and Minderer, Matthias and Heigold, Georg and Gelly, Sylvain and others},
  journal={arXiv preprint arXiv:2010.11929},
  year={2020}
}

@inproceedings{deng2022anomaly,
  title={Anomaly detection via reverse distillation from one-class embedding},
  author={Deng, Hanqiu and Li, Xingyu},
  booktitle={Proceedings of the IEEE/CVF conference on computer vision and pattern recognition},
  pages={9737--9746},
  year={2022}
}

@inproceedings{tien2023revisiting,
  title={Revisiting reverse distillation for anomaly detection},
  author={Tien, Tran Dinh and Nguyen, Anh Tuan and Tran, Nguyen Hoang and Huy, Ta Duc and Duong, Soan and Nguyen, Chanh D Tr and Truong, Steven QH},
  booktitle={Proceedings of the IEEE/CVF conference on computer vision and pattern recognition},
  pages={24511--24520},
  year={2023}
}

@article{ren2025foundation,
  title={Foundation models for anomaly detection: Vision and challenges},
  author={Ren, Jing and Tang, Tao and Jia, Hong and Xu, Ziqi and Fayek, Haytham and Li, Xiaodong and Ma, Suyu and Xu, Xiwei and Xia, Feng},
  journal={AI Magazine},
  volume={46},
  number={4},
  pages={e70045},
  year={2025},
  publisher={Wiley Online Library}
}

@article{ammar2025foundation,
  title={Foundation models and Transformers for anomaly detection: A survey},
  author={Ammar, Mouin Ben and Mendoza, Arturo and Belkhir, Nacim and Manzanera, Antoine and Franchi, Gianni},
  journal={Information Fusion},
  pages={103517},
  year={2025},
  publisher={Elsevier}
}

@inproceedings{he2022masked,
  title={Masked autoencoders are scalable vision learners},
  author={He, Kaiming and Chen, Xinlei and Xie, Saining and Li, Yanghao and Doll{\'a}r, Piotr and Girshick, Ross},
  booktitle={Proceedings of the IEEE/CVF conference on computer vision and pattern recognition},
  pages={16000--16009},
  year={2022}
}

@inproceedings{caron2021emerging,
  title={Emerging properties in self-supervised vision transformers},
  author={Caron, Mathilde and Touvron, Hugo and Misra, Ishan and J{\'e}gou, Herv{\'e} and Mairal, Julien and Bojanowski, Piotr and Joulin, Armand},
  booktitle={Proceedings of the IEEE/CVF international conference on computer vision},
  pages={9650--9660},
  year={2021}
}

@inproceedings{radford2021learning,
  title={Learning transferable visual models from natural language supervision},
  author={Radford, Alec and Kim, Jong Wook and Hallacy, Chris and Ramesh, Aditya and Goh, Gabriel and Agarwal, Sandhini and Sastry, Girish and Askell, Amanda and Mishkin, Pamela and Clark, Jack and others},
  booktitle={International conference on machine learning},
  pages={8748--8763},
  year={2021},
  organization={PmLR}
}

@article{fort2021exploring,
  title={Exploring the limits of out-of-distribution detection},
  author={Fort, Stanislav and Ren, Jie and Lakshminarayanan, Balaji},
  journal={Advances in neural information processing systems},
  volume={34},
  pages={7068--7081},
  year={2021}
}

@article{meinke2022provably,
  title={Provably ad-versarially robust detection of out-of-distribution data},
  author={Meinke, Alexander and Bitterwolf, J and Hein, M},
  journal={Advances in Neural Information Processing Systems (Ne urIPS)},
  volume={35},
  year={2022}
}

@article{bitterwolf2020certifiably,
  title={Certifiably adversarially robust detection of out-of-distribution data},
  author={Bitterwolf, Julian and Meinke, Alexander and Hein, Matthias},
  journal={Advances in Neural Information Processing Systems},
  volume={33},
  pages={16085--16095},
  year={2020}
}

@inproceedings{croce2020reliable,
  title={Reliable evaluation of adversarial robustness with an ensemble of diverse parameter-free attacks},
  author={Croce, Francesco and Hein, Matthias},
  booktitle={International conference on machine learning},
  pages={2206--2216},
  year={2020},
  organization={PMLR}
}

@inproceedings{bachmann2022multimae,
  title={Multimae: Multi-modal multi-task masked autoencoders},
  author={Bachmann, Roman and Mizrahi, David and Atanov, Andrei and Zamir, Amir},
  booktitle={European conference on computer vision},
  pages={348--367},
  year={2022},
  organization={Springer}
}
\end{document}


\maketitle


\section{Augmentation Details}
\label{sec:supp_augmentation}

This section documents the data augmentations used at each stage of
our pipeline. We deliberately keep augmentation minimal as the
few-shot setting motivates leveraging the small support set
faithfully rather than synthesizing additional training distributions
that could obscure the contribution of the DistanceProbe and
PatchShift\,+\,FeaturePurifier defense.

\subsection{Per-Object Rotation Augmentation}
\label{sec:supp_rotation}

For categories in which nominal training images appear in arbitrary
orientations only, we augment the support set, the
DistanceProbe training pool, and the FeaturePurifier training pool
with rotations of $\{45^\circ, 90^\circ, 135^\circ, 180^\circ,
225^\circ, 270^\circ, 315^\circ\}$ applied via affine warp around the
image center, yielding eight oriented views per original image
(including the unrotated original). Following
AnomalyDINO~\cite{damm2025anomalydino}, we adopt the \emph{informed}
preprocessing policy on MVTec-AD~\cite{bergmann2019mvtec}: rotation
is enabled only for \emph{hazelnut} and \emph{screw}, the two MVTec
categories whose nominal training images exhibit free rotation, and
disabled for the remaining $13$ categories whose training images are
canonically oriented. For VisA~\cite{zou2022spot} and
MPDD~\cite{jezek2021deep}, the dataset-default policies are mirrored
identically. The same per-object rotation policy is applied
consistently at three sites: (i)~when building the support memory
bank $\mathcal{M}$, (ii)~when extracting features for DistanceProbe
training, and (iii)~when extracting clean features and generating
adversarial counterparts for FeaturePurifier training. This ensures
that the memory bank, the probe, and the purifier all see the same
distribution of orientations.

\subsection{Per-Object Background Masking}
\label{sec:supp_masking}

For categories whose nominal images contain large near-uniform
background regions, we apply a PCA-based foreground mask
following~\cite{damm2025anomalydino} by computing the first
principal component of the support memory bank features and drop
patch tokens whose first-PC magnitude falls below a fixed threshold
of $10$. On MVTec-AD, masking is enabled for \emph{capsule},
\emph{hazelnut}, \emph{pill}, \emph{screw}, and \emph{toothbrush}; it
is enabled for all categories on VisA and MPDD, where background
heterogeneity is consistent across the dataset. Masking is technically
not augmentation but per-image preprocessing and we list it here because
the masking and rotation policies share the same per-object
configuration and are jointly determined by the choice of
preprocessing regime (\emph{informed} or \emph{agnostic}).

\subsection{Gaussian Noise Augmentation for the FeaturePurifier}
\label{sec:supp_noise}

The FeaturePurifier is trained on (clean, perturbed) feature pairs
extracted from the $k$ support images. In addition to the adversarial
pairs generated by the DistanceProbe-driven PGD attack, we augment the perturbed half of each
training pair with $L_{\mathrm{noise}} = 4$ Gaussian-noise variants
per support image. For each variant, a noise tensor
$\eta \sim \mathcal{N}(0, \sigma_{\mathrm{noise}}^2 I)$ is sampled
with $\sigma_{\mathrm{noise}} = 0.03$ in pixel $[0, 1]$ space, scaled
per channel by the ImageNet-normalization standard deviation $1/\sigma$
to match the backbone's input space, added to the normalized image,
and clamped to the valid normalized range
$[(0 - \mu)/\sigma, (1 - \mu)/\sigma]$. The resulting noised tensor
is passed through the frozen DINOv2 backbone, and the masked patch
features are paired with their clean counterparts as additional
purifier training examples. This complements the adversarial pairs
by exposing the purifier to a broader class of input perturbations,
encouraging generalization beyond the specific PGD attack used to
generate the adversarial pairs.

\subsection{Augmentations Not Used}
\label{sec:supp_no_aug}

We deliberately do not apply common image augmentations such as
color jitter, horizontal or vertical flipping, random cropping,
resized cropping, scale jitter, CutPaste~\cite{li2021cutpaste}, or
synthetic anomaly generation. Test images receive only deterministic
resizing to a smaller edge of $448$ pixels and center-cropping to
the largest multiple of the patch size $P = 14$. The random pixel
shifts introduced by PatchShift at inference time are part of the
defense pipeline, not training-time augmentation and they are sampled
fresh at each test forward pass and applied to both clean and
adversarial inputs alike during evaluation.

\section{Additional Ablation: Backbone Variant}
\label{sec:supp_backbone}

We evaluate our defense across three DINOv2 backbone
variants~\cite{oquab2023dinov2}: ViT-S/14 (22M parameters,
$D = 384$), ViT-B/14 (86M, $D = 768$), and ViT-L/14 (304M, $D = 1024$),
on MVTec-AD at $k = 4$ shots under PGD-20 at $\epsilon = 8/255$.
Table~\ref{tab:supp_backbone} reports the results. Probe and purifier hidden dimensions were scaled to match each backbone's feature dimension ($h=D$) and other hyperparameters were kept identical. ViT-S/14 yields the
strongest robust performance on both metrics, with ViT-B/14 and
ViT-L/14 trailing by $6$--$13$ points image-level AUROC and
$13$--$18$ points PRO. This mirrors the observation by
AnomalyDINO~\cite{damm2025anomalydino} that ViT-S/14 features are the
strongest choice for $k$-NN-based FSAD even in the clean setting, and
the gap widens under adversarial perturbation, consistent with
larger backbones amplifying small input deviations into larger
feature-space displacements through their deeper layer stacks. We
adopt ViT-S/14 throughout the main paper. 

\begin{table}[h]
\centering
\caption{Robust image-level AUROC and pixel-level PRO (\%) on
MVTec-AD across DINOv2 backbone variants. $k = 4$ shots,
PGD-20 at $\epsilon = 8/255$, averaged across $3$ random seeds.}
\label{tab:supp_backbone}
\setlength{\tabcolsep}{6pt}
\small
\begin{tabular}{lcc}
\toprule
\textbf{Backbone} & \textbf{Image AUROC} & \textbf{PRO} \\
\midrule
ViT-S/14        & \textbf{76.2} & \textbf{64.4} \\
ViT-B/14        & 68.0          & 51.3 \\
ViT-L/14        & 69.8          & 46.1 \\
\bottomrule
\end{tabular}
\end{table}

\begin{table}[h]
\centering
\caption{Robust image-level AUROC and pixel-level PRO (\%) under
$\ell_2$-PGD-20 attack at $k = 4$ shots,
$\epsilon_{\ell_2} = 64/255$.}
\label{tab:supp_l2_pgd}
\setlength{\tabcolsep}{6pt}
\small
\begin{tabular}{lccc}
\toprule
\textbf{Metric} & \textbf{MVTec-AD} & \textbf{VisA} & \textbf{MPDD} \\
\midrule
Image-level AUROC & \textbf{93.6} & \textbf{84.3} & \textbf{73.3} \\
PRO               & \textbf{89.7} & \textbf{89.8} & \textbf{83.8} \\
\bottomrule
\end{tabular}
\end{table}

\section{Additional Results: $L_2$ PGD-20 Attack}

\label{sec:supp_l2_pgd}

While our main experiments report $\ell_\infty$-PGD as the primary
threat model, we additionally evaluate our defense against
$\ell_2$-bounded PGD-20 attacks to characterize robustness across the
standard $p$-norm family.
Table~\ref{tab:supp_l2_pgd} reports robust image-level AUROC and
pixel-level PRO at $k = 4$ shots on MVTec-AD, VisA, and MPDD against
$\ell_2$-PGD-20 at $\epsilon_{\ell_2} = 64/255$. Both
metrics remain close to their clean-defended values across all three
datasets, with image-level AUROC reaching $93.6/84.3/73.3\%$ and PRO
reaching $89.7/89.8/83.8\%$ on MVTec-AD/VisA/MPDD respectively. The
$\ell_2$-bounded attacker is structurally weaker than the
$\ell_\infty$ attacker at comparable nominal budgets because the
$\ell_2$ constraint spreads the perturbation magnitude across all
pixels rather than allowing per-pixel saturation; the smaller
per-pixel deviations translate to smaller feature-space displacements
and a less destructive attack. The defense correspondingly degrades
less under $\ell_2$ than under $\ell_\infty$ at $\epsilon = 8/255$, consistent with this difference
in threat-model strength rather than indicating $\ell_2$-specific
robustness.

\section{Per-Category Results}
In this section we present the per-category results on all three datasets (MVTec\,AD, VisA, MPDD) across all four few-shot configurations ($k\!\in\!\{1,2,4,8\}$) under both clean-defended evaluation and robust evaluation against a PGD-20 adversarial attack with $\epsilon\!=\!8/255$ as detailed in Tables~\ref{tab:mvtec_1shot}, \ref{tab:mvtec_2shot}, \ref{tab:mvtec_4shot}, \ref{tab:mvtec_8shot}, \ref{tab:visa_1shot}, \ref{tab:visa_2shot}, \ref{tab:visa_4shot}, \ref{tab:visa_8shot}, \ref{tab:mpdd_1shot}, \ref{tab:mpdd_2shot}, \ref{tab:mpdd_4shot}, and \ref{tab:mpdd_8shot}. For every category we report classification AUROC, segmentation AUROC, and per-region overlap (PRO) as mean$\pm$std across three seeds.

\begin{table}[h]
\centering
\caption{Per-category results on MVTec (1-shot). Each cell reports mean$\pm$std across 3 seeds for clean-defended and PGD-20 ($\epsilon=8/255$) robust evaluation. All values are percentages.}
\setlength{\tabcolsep}{4pt}
\resizebox{\linewidth}{!}{%
\begin{tabular}{lcccccc}
\toprule
 & \multicolumn{2}{c}{Image\ AUROC} & \multicolumn{2}{c}{Pixel\ AUROC} & \multicolumn{2}{c}{PRO} \\
\cmidrule(lr){2-3} \cmidrule(lr){4-5} \cmidrule(lr){6-7}
Object & Clean & Robust & Clean & Robust & Clean & Robust \\
\midrule
bottle & 100.00\textsubscript{\scriptsize $\pm$0.00} & 89.92\textsubscript{\scriptsize $\pm$8.62} & 99.15\textsubscript{\scriptsize $\pm$0.04} & 88.79\textsubscript{\scriptsize $\pm$2.61} & 95.54\textsubscript{\scriptsize $\pm$0.35} & 68.69\textsubscript{\scriptsize $\pm$3.92} \\
cable & 94.08\textsubscript{\scriptsize $\pm$0.76} & 56.72\textsubscript{\scriptsize $\pm$6.92} & 94.29\textsubscript{\scriptsize $\pm$0.63} & 66.05\textsubscript{\scriptsize $\pm$1.54} & 87.48\textsubscript{\scriptsize $\pm$0.78} & 40.11\textsubscript{\scriptsize $\pm$1.27} \\
capsule & 87.37\textsubscript{\scriptsize $\pm$8.30} & 48.44\textsubscript{\scriptsize $\pm$6.10} & 98.49\textsubscript{\scriptsize $\pm$0.30} & 89.77\textsubscript{\scriptsize $\pm$1.58} & 94.75\textsubscript{\scriptsize $\pm$0.77} & 63.18\textsubscript{\scriptsize $\pm$4.47} \\
carpet & 100.00\textsubscript{\scriptsize $\pm$0.00} & 96.33\textsubscript{\scriptsize $\pm$2.02} & 99.47\textsubscript{\scriptsize $\pm$0.01} & 96.79\textsubscript{\scriptsize $\pm$1.06} & 97.87\textsubscript{\scriptsize $\pm$0.15} & 85.93\textsubscript{\scriptsize $\pm$5.13} \\
grid & 99.67\textsubscript{\scriptsize $\pm$0.17} & 96.69\textsubscript{\scriptsize $\pm$0.54} & 98.93\textsubscript{\scriptsize $\pm$0.06} & 95.35\textsubscript{\scriptsize $\pm$2.04} & 95.26\textsubscript{\scriptsize $\pm$0.34} & 86.69\textsubscript{\scriptsize $\pm$5.24} \\
hazelnut & 96.18\textsubscript{\scriptsize $\pm$2.06} & 32.92\textsubscript{\scriptsize $\pm$4.73} & 99.38\textsubscript{\scriptsize $\pm$0.09} & 88.77\textsubscript{\scriptsize $\pm$3.40} & 95.57\textsubscript{\scriptsize $\pm$0.43} & 57.62\textsubscript{\scriptsize $\pm$2.05} \\
leather & 100.00\textsubscript{\scriptsize $\pm$0.00} & 99.71\textsubscript{\scriptsize $\pm$0.14} & 98.74\textsubscript{\scriptsize $\pm$0.08} & 97.86\textsubscript{\scriptsize $\pm$0.03} & 96.78\textsubscript{\scriptsize $\pm$0.55} & 92.00\textsubscript{\scriptsize $\pm$1.66} \\
metal\_nut & 100.00\textsubscript{\scriptsize $\pm$0.00} & 71.13\textsubscript{\scriptsize $\pm$4.54} & 96.81\textsubscript{\scriptsize $\pm$0.37} & 27.74\textsubscript{\scriptsize $\pm$11.09} & 93.73\textsubscript{\scriptsize $\pm$0.29} & 23.36\textsubscript{\scriptsize $\pm$10.63} \\
pill & 96.84\textsubscript{\scriptsize $\pm$0.85} & 42.45\textsubscript{\scriptsize $\pm$23.77} & 98.32\textsubscript{\scriptsize $\pm$0.27} & 65.08\textsubscript{\scriptsize $\pm$4.38} & 96.88\textsubscript{\scriptsize $\pm$0.36} & 30.76\textsubscript{\scriptsize $\pm$8.81} \\
screw & 82.95\textsubscript{\scriptsize $\pm$3.45} & 26.95\textsubscript{\scriptsize $\pm$6.91} & 97.54\textsubscript{\scriptsize $\pm$0.18} & 84.78\textsubscript{\scriptsize $\pm$4.22} & 88.73\textsubscript{\scriptsize $\pm$0.85} & 50.16\textsubscript{\scriptsize $\pm$12.54} \\
tile & 100.00\textsubscript{\scriptsize $\pm$0.00} & 86.03\textsubscript{\scriptsize $\pm$7.51} & 96.10\textsubscript{\scriptsize $\pm$0.19} & 76.76\textsubscript{\scriptsize $\pm$4.50} & 88.24\textsubscript{\scriptsize $\pm$0.41} & 47.48\textsubscript{\scriptsize $\pm$2.52} \\
toothbrush & 96.30\textsubscript{\scriptsize $\pm$0.89} & 57.78\textsubscript{\scriptsize $\pm$12.97} & 99.07\textsubscript{\scriptsize $\pm$0.08} & 95.07\textsubscript{\scriptsize $\pm$1.11} & 89.18\textsubscript{\scriptsize $\pm$0.38} & 70.40\textsubscript{\scriptsize $\pm$3.57} \\
transistor & 94.24\textsubscript{\scriptsize $\pm$1.18} & 52.21\textsubscript{\scriptsize $\pm$10.65} & 89.28\textsubscript{\scriptsize $\pm$0.35} & 29.49\textsubscript{\scriptsize $\pm$8.88} & 71.91\textsubscript{\scriptsize $\pm$1.08} & 11.41\textsubscript{\scriptsize $\pm$2.88} \\
wood & 99.82\textsubscript{\scriptsize $\pm$0.09} & 44.39\textsubscript{\scriptsize $\pm$5.40} & 95.89\textsubscript{\scriptsize $\pm$0.42} & 66.78\textsubscript{\scriptsize $\pm$1.79} & 90.33\textsubscript{\scriptsize $\pm$1.05} & 32.56\textsubscript{\scriptsize $\pm$3.77} \\
zipper & 98.44\textsubscript{\scriptsize $\pm$0.24} & 74.25\textsubscript{\scriptsize $\pm$10.23} & 95.11\textsubscript{\scriptsize $\pm$0.38} & 45.67\textsubscript{\scriptsize $\pm$1.33} & 87.07\textsubscript{\scriptsize $\pm$0.88} & 40.14\textsubscript{\scriptsize $\pm$2.02} \\
\midrule
\textbf{Average} & \textbf{96.39\textsubscript{\scriptsize $\pm$0.73}} & \textbf{65.06\textsubscript{\scriptsize $\pm$0.46}} & \textbf{97.10\textsubscript{\scriptsize $\pm$0.07}} & \textbf{74.32\textsubscript{\scriptsize $\pm$0.49}} & \textbf{91.29\textsubscript{\scriptsize $\pm$0.14}} & \textbf{53.37\textsubscript{\scriptsize $\pm$0.82}} \\
\bottomrule
\end{tabular}}
\label{tab:mvtec_1shot}
\end{table}

\begin{table}[H]
\centering
\caption{Per-category results on MVTec (2-shot). Each cell reports mean$\pm$std across 3 seeds for clean-defended and PGD-20 ($\epsilon=8/255$) robust evaluation. All values are percentages.}
\setlength{\tabcolsep}{4pt}
\resizebox{\linewidth}{!}{%
\begin{tabular}{lcccccc}
\toprule
 & \multicolumn{2}{c}{Image\ AUROC} & \multicolumn{2}{c}{Pixel\ AUROC} & \multicolumn{2}{c}{PRO} \\
\cmidrule(lr){2-3} \cmidrule(lr){4-5} \cmidrule(lr){6-7}
Object & Clean & Robust & Clean & Robust & Clean & Robust \\
\midrule
bottle & 100.00\textsubscript{\scriptsize $\pm$0.00} & 87.09\textsubscript{\scriptsize $\pm$15.23} & 99.20\textsubscript{\scriptsize $\pm$0.02} & 89.94\textsubscript{\scriptsize $\pm$4.35} & 95.53\textsubscript{\scriptsize $\pm$0.33} & 67.58\textsubscript{\scriptsize $\pm$11.93} \\
cable & 93.91\textsubscript{\scriptsize $\pm$1.21} & 55.94\textsubscript{\scriptsize $\pm$3.69} & 94.75\textsubscript{\scriptsize $\pm$0.65} & 66.25\textsubscript{\scriptsize $\pm$3.84} & 87.76\textsubscript{\scriptsize $\pm$0.74} & 40.28\textsubscript{\scriptsize $\pm$6.31} \\
capsule & 88.54\textsubscript{\scriptsize $\pm$7.70} & 59.07\textsubscript{\scriptsize $\pm$5.72} & 98.66\textsubscript{\scriptsize $\pm$0.23} & 89.57\textsubscript{\scriptsize $\pm$1.65} & 95.15\textsubscript{\scriptsize $\pm$1.03} & 61.94\textsubscript{\scriptsize $\pm$4.48} \\
carpet & 100.00\textsubscript{\scriptsize $\pm$0.00} & 97.11\textsubscript{\scriptsize $\pm$1.07} & 99.53\textsubscript{\scriptsize $\pm$0.04} & 97.01\textsubscript{\scriptsize $\pm$0.34} & 97.88\textsubscript{\scriptsize $\pm$0.10} & 87.21\textsubscript{\scriptsize $\pm$0.55} \\
grid & 99.78\textsubscript{\scriptsize $\pm$0.21} & 98.44\textsubscript{\scriptsize $\pm$0.97} & 98.94\textsubscript{\scriptsize $\pm$0.10} & 96.70\textsubscript{\scriptsize $\pm$1.09} & 95.35\textsubscript{\scriptsize $\pm$0.50} & 88.70\textsubscript{\scriptsize $\pm$3.09} \\
hazelnut & 99.45\textsubscript{\scriptsize $\pm$0.95} & 45.02\textsubscript{\scriptsize $\pm$19.43} & 99.50\textsubscript{\scriptsize $\pm$0.01} & 90.32\textsubscript{\scriptsize $\pm$2.32} & 95.98\textsubscript{\scriptsize $\pm$0.56} & 67.59\textsubscript{\scriptsize $\pm$7.19} \\
leather & 100.00\textsubscript{\scriptsize $\pm$0.00} & 99.28\textsubscript{\scriptsize $\pm$0.37} & 98.86\textsubscript{\scriptsize $\pm$0.26} & 97.51\textsubscript{\scriptsize $\pm$0.39} & 97.06\textsubscript{\scriptsize $\pm$0.25} & 91.06\textsubscript{\scriptsize $\pm$1.17} \\
metal\_nut & 100.00\textsubscript{\scriptsize $\pm$0.00} & 63.93\textsubscript{\scriptsize $\pm$10.41} & 96.97\textsubscript{\scriptsize $\pm$0.12} & 39.33\textsubscript{\scriptsize $\pm$1.64} & 94.01\textsubscript{\scriptsize $\pm$0.39} & 40.23\textsubscript{\scriptsize $\pm$3.61} \\
pill & 96.82\textsubscript{\scriptsize $\pm$1.93} & 69.39\textsubscript{\scriptsize $\pm$6.04} & 98.36\textsubscript{\scriptsize $\pm$0.22} & 67.51\textsubscript{\scriptsize $\pm$1.00} & 96.98\textsubscript{\scriptsize $\pm$0.36} & 34.24\textsubscript{\scriptsize $\pm$4.07} \\
screw & 87.16\textsubscript{\scriptsize $\pm$2.36} & 34.80\textsubscript{\scriptsize $\pm$2.57} & 97.87\textsubscript{\scriptsize $\pm$0.21} & 88.31\textsubscript{\scriptsize $\pm$1.68} & 90.06\textsubscript{\scriptsize $\pm$0.69} & 59.96\textsubscript{\scriptsize $\pm$5.49} \\
tile & 100.00\textsubscript{\scriptsize $\pm$0.00} & 95.59\textsubscript{\scriptsize $\pm$3.61} & 95.39\textsubscript{\scriptsize $\pm$0.42} & 83.21\textsubscript{\scriptsize $\pm$4.93} & 86.70\textsubscript{\scriptsize $\pm$1.41} & 54.45\textsubscript{\scriptsize $\pm$8.09} \\
toothbrush & 96.48\textsubscript{\scriptsize $\pm$1.53} & 71.76\textsubscript{\scriptsize $\pm$7.03} & 99.04\textsubscript{\scriptsize $\pm$0.08} & 96.33\textsubscript{\scriptsize $\pm$0.61} & 89.12\textsubscript{\scriptsize $\pm$0.92} & 75.08\textsubscript{\scriptsize $\pm$0.58} \\
transistor & 91.03\textsubscript{\scriptsize $\pm$7.66} & 46.00\textsubscript{\scriptsize $\pm$8.23} & 87.38\textsubscript{\scriptsize $\pm$5.94} & 33.06\textsubscript{\scriptsize $\pm$4.43} & 70.44\textsubscript{\scriptsize $\pm$5.67} & 13.20\textsubscript{\scriptsize $\pm$4.46} \\
wood & 99.53\textsubscript{\scriptsize $\pm$0.51} & 83.10\textsubscript{\scriptsize $\pm$0.35} & 95.88\textsubscript{\scriptsize $\pm$0.16} & 79.65\textsubscript{\scriptsize $\pm$4.99} & 89.49\textsubscript{\scriptsize $\pm$0.17} & 62.15\textsubscript{\scriptsize $\pm$9.74} \\
zipper & 99.05\textsubscript{\scriptsize $\pm$0.18} & 87.25\textsubscript{\scriptsize $\pm$4.36} & 95.73\textsubscript{\scriptsize $\pm$0.63} & 56.56\textsubscript{\scriptsize $\pm$3.27} & 88.40\textsubscript{\scriptsize $\pm$1.48} & 47.50\textsubscript{\scriptsize $\pm$5.11} \\
\midrule
\textbf{Average} & \textbf{96.78\textsubscript{\scriptsize $\pm$1.02}} & \textbf{72.92\textsubscript{\scriptsize $\pm$1.12}} & \textbf{97.07\textsubscript{\scriptsize $\pm$0.41}} & \textbf{78.08\textsubscript{\scriptsize $\pm$0.71}} & \textbf{91.33\textsubscript{\scriptsize $\pm$0.43}} & \textbf{59.41\textsubscript{\scriptsize $\pm$0.46}} \\
\bottomrule
\end{tabular}}
\label{tab:mvtec_2shot}
\end{table}

\begin{table}[H]
\centering
\caption{Per-category results on MVTec (4-shot). Each cell reports mean$\pm$std across 3 seeds for clean-defended and PGD-20 ($\epsilon=8/255$) robust evaluation. All values are percentages.}
\setlength{\tabcolsep}{4pt}
\resizebox{\linewidth}{!}{%
\begin{tabular}{lcccccc}
\toprule
 & \multicolumn{2}{c}{Image\ AUROC} & \multicolumn{2}{c}{Pixel\ AUROC} & \multicolumn{2}{c}{PRO} \\
\cmidrule(lr){2-3} \cmidrule(lr){4-5} \cmidrule(lr){6-7}
Object & Clean & Robust & Clean & Robust & Clean & Robust \\
\midrule
bottle & 100.00\textsubscript{\scriptsize $\pm$0.00} & 97.80\textsubscript{\scriptsize $\pm$0.12} & 99.22\textsubscript{\scriptsize $\pm$0.00} & 92.63\textsubscript{\scriptsize $\pm$1.54} & 95.33\textsubscript{\scriptsize $\pm$0.41} & 75.09\textsubscript{\scriptsize $\pm$1.54} \\
cable & 95.91\textsubscript{\scriptsize $\pm$1.39} & 76.21\textsubscript{\scriptsize $\pm$9.77} & 95.30\textsubscript{\scriptsize $\pm$0.51} & 76.37\textsubscript{\scriptsize $\pm$3.52} & 88.85\textsubscript{\scriptsize $\pm$0.72} & 58.54\textsubscript{\scriptsize $\pm$3.39} \\
capsule & 95.32\textsubscript{\scriptsize $\pm$1.04} & 70.58\textsubscript{\scriptsize $\pm$7.62} & 98.88\textsubscript{\scriptsize $\pm$0.03} & 90.48\textsubscript{\scriptsize $\pm$0.26} & 95.97\textsubscript{\scriptsize $\pm$0.10} & 67.26\textsubscript{\scriptsize $\pm$2.92} \\
carpet & 100.00\textsubscript{\scriptsize $\pm$0.00} & 98.45\textsubscript{\scriptsize $\pm$0.52} & 99.51\textsubscript{\scriptsize $\pm$0.08} & 97.94\textsubscript{\scriptsize $\pm$0.43} & 97.85\textsubscript{\scriptsize $\pm$0.16} & 91.61\textsubscript{\scriptsize $\pm$0.73} \\
grid & 99.94\textsubscript{\scriptsize $\pm$0.10} & 99.30\textsubscript{\scriptsize $\pm$0.41} & 98.97\textsubscript{\scriptsize $\pm$0.13} & 97.16\textsubscript{\scriptsize $\pm$0.83} & 95.52\textsubscript{\scriptsize $\pm$0.67} & 91.16\textsubscript{\scriptsize $\pm$0.99} \\
hazelnut & 99.75\textsubscript{\scriptsize $\pm$0.34} & 54.00\textsubscript{\scriptsize $\pm$16.59} & 99.52\textsubscript{\scriptsize $\pm$0.01} & 93.93\textsubscript{\scriptsize $\pm$1.33} & 96.57\textsubscript{\scriptsize $\pm$0.09} & 73.30\textsubscript{\scriptsize $\pm$7.90} \\
leather & 100.00\textsubscript{\scriptsize $\pm$0.00} & 99.91\textsubscript{\scriptsize $\pm$0.10} & 98.86\textsubscript{\scriptsize $\pm$0.15} & 98.26\textsubscript{\scriptsize $\pm$0.14} & 97.13\textsubscript{\scriptsize $\pm$0.45} & 94.04\textsubscript{\scriptsize $\pm$0.83} \\
metal\_nut & 100.00\textsubscript{\scriptsize $\pm$0.00} & 71.34\textsubscript{\scriptsize $\pm$7.22} & 97.44\textsubscript{\scriptsize $\pm$0.11} & 50.95\textsubscript{\scriptsize $\pm$17.48} & 94.77\textsubscript{\scriptsize $\pm$0.08} & 46.50\textsubscript{\scriptsize $\pm$12.56} \\
pill & 97.88\textsubscript{\scriptsize $\pm$0.32} & 58.44\textsubscript{\scriptsize $\pm$6.25} & 98.45\textsubscript{\scriptsize $\pm$0.24} & 69.01\textsubscript{\scriptsize $\pm$5.13} & 97.27\textsubscript{\scriptsize $\pm$0.10} & 37.35\textsubscript{\scriptsize $\pm$6.59} \\
screw & 87.32\textsubscript{\scriptsize $\pm$4.21} & 24.12\textsubscript{\scriptsize $\pm$4.58} & 98.15\textsubscript{\scriptsize $\pm$0.64} & 88.62\textsubscript{\scriptsize $\pm$1.09} & 90.89\textsubscript{\scriptsize $\pm$2.58} & 61.92\textsubscript{\scriptsize $\pm$1.80} \\
tile & 100.00\textsubscript{\scriptsize $\pm$0.00} & 97.39\textsubscript{\scriptsize $\pm$3.97} & 95.64\textsubscript{\scriptsize $\pm$0.91} & 86.47\textsubscript{\scriptsize $\pm$0.78} & 87.37\textsubscript{\scriptsize $\pm$1.88} & 63.06\textsubscript{\scriptsize $\pm$0.53} \\
toothbrush & 96.67\textsubscript{\scriptsize $\pm$1.92} & 71.11\textsubscript{\scriptsize $\pm$7.40} & 99.00\textsubscript{\scriptsize $\pm$0.03} & 95.94\textsubscript{\scriptsize $\pm$1.07} & 89.32\textsubscript{\scriptsize $\pm$0.86} & 75.09\textsubscript{\scriptsize $\pm$5.12} \\
transistor & 94.28\textsubscript{\scriptsize $\pm$1.85} & 57.90\textsubscript{\scriptsize $\pm$12.49} & 89.66\textsubscript{\scriptsize $\pm$2.05} & 39.64\textsubscript{\scriptsize $\pm$6.90} & 73.40\textsubscript{\scriptsize $\pm$1.86} & 19.43\textsubscript{\scriptsize $\pm$1.50} \\
wood & 99.06\textsubscript{\scriptsize $\pm$0.79} & 79.33\textsubscript{\scriptsize $\pm$11.53} & 95.65\textsubscript{\scriptsize $\pm$0.32} & 80.08\textsubscript{\scriptsize $\pm$4.54} & 89.52\textsubscript{\scriptsize $\pm$1.11} & 62.96\textsubscript{\scriptsize $\pm$3.75} \\
zipper & 99.22\textsubscript{\scriptsize $\pm$0.15} & 86.44\textsubscript{\scriptsize $\pm$4.78} & 95.88\textsubscript{\scriptsize $\pm$0.18} & 58.24\textsubscript{\scriptsize $\pm$5.54} & 88.68\textsubscript{\scriptsize $\pm$0.24} & 48.95\textsubscript{\scriptsize $\pm$2.45} \\
\midrule
\textbf{Average} & \textbf{97.69\textsubscript{\scriptsize $\pm$0.22}} & \textbf{76.15\textsubscript{\scriptsize $\pm$1.92}} & \textbf{97.34\textsubscript{\scriptsize $\pm$0.07}} & \textbf{81.05\textsubscript{\scriptsize $\pm$1.80}} & \textbf{91.90\textsubscript{\scriptsize $\pm$0.23}} & \textbf{64.42\textsubscript{\scriptsize $\pm$1.30}} \\
\bottomrule
\end{tabular}}
\label{tab:mvtec_4shot}
\end{table}

\begin{table}[H]
\centering
\caption{Per-category results on MVTec (8-shot). Each cell reports mean$\pm$std across 3 seeds for clean-defended and PGD-20 ($\epsilon=8/255$) robust evaluation. All values are percentages.}
\setlength{\tabcolsep}{4pt}
\resizebox{\linewidth}{!}{%
\begin{tabular}{lcccccc}
\toprule
 & \multicolumn{2}{c}{Image\ AUROC} & \multicolumn{2}{c}{Pixel\ AUROC} & \multicolumn{2}{c}{PRO} \\
\cmidrule(lr){2-3} \cmidrule(lr){4-5} \cmidrule(lr){6-7}
Object & Clean & Robust & Clean & Robust & Clean & Robust \\
\midrule
bottle & 100.00\textsubscript{\scriptsize $\pm$0.00} & 97.75\textsubscript{\scriptsize $\pm$1.11} & 99.23\textsubscript{\scriptsize $\pm$0.02} & 94.12\textsubscript{\scriptsize $\pm$1.69} & 95.14\textsubscript{\scriptsize $\pm$0.40} & 80.48\textsubscript{\scriptsize $\pm$3.68} \\
cable & 96.89\textsubscript{\scriptsize $\pm$0.39} & 82.23\textsubscript{\scriptsize $\pm$11.43} & 95.51\textsubscript{\scriptsize $\pm$0.05} & 79.40\textsubscript{\scriptsize $\pm$1.49} & 89.34\textsubscript{\scriptsize $\pm$0.11} & 63.13\textsubscript{\scriptsize $\pm$3.30} \\
capsule & 94.32\textsubscript{\scriptsize $\pm$2.62} & 73.94\textsubscript{\scriptsize $\pm$10.16} & 98.95\textsubscript{\scriptsize $\pm$0.03} & 90.14\textsubscript{\scriptsize $\pm$1.31} & 96.00\textsubscript{\scriptsize $\pm$0.16} & 66.10\textsubscript{\scriptsize $\pm$4.20} \\
carpet & 100.00\textsubscript{\scriptsize $\pm$0.00} & 96.50\textsubscript{\scriptsize $\pm$0.87} & 99.55\textsubscript{\scriptsize $\pm$0.04} & 96.85\textsubscript{\scriptsize $\pm$0.86} & 97.93\textsubscript{\scriptsize $\pm$0.03} & 85.76\textsubscript{\scriptsize $\pm$2.31} \\
grid & 100.00\textsubscript{\scriptsize $\pm$0.00} & 98.97\textsubscript{\scriptsize $\pm$0.79} & 98.97\textsubscript{\scriptsize $\pm$0.08} & 97.42\textsubscript{\scriptsize $\pm$0.86} & 95.70\textsubscript{\scriptsize $\pm$0.24} & 90.13\textsubscript{\scriptsize $\pm$3.04} \\
hazelnut & 100.00\textsubscript{\scriptsize $\pm$0.00} & 72.24\textsubscript{\scriptsize $\pm$3.59} & 99.55\textsubscript{\scriptsize $\pm$0.03} & 96.09\textsubscript{\scriptsize $\pm$0.13} & 96.53\textsubscript{\scriptsize $\pm$0.29} & 77.45\textsubscript{\scriptsize $\pm$2.01} \\
leather & 100.00\textsubscript{\scriptsize $\pm$0.00} & 99.90\textsubscript{\scriptsize $\pm$0.09} & 98.92\textsubscript{\scriptsize $\pm$0.13} & 98.32\textsubscript{\scriptsize $\pm$0.18} & 97.30\textsubscript{\scriptsize $\pm$0.27} & 94.24\textsubscript{\scriptsize $\pm$1.86} \\
metal\_nut & 100.00\textsubscript{\scriptsize $\pm$0.00} & 79.65\textsubscript{\scriptsize $\pm$2.32} & 97.45\textsubscript{\scriptsize $\pm$0.24} & 60.37\textsubscript{\scriptsize $\pm$11.30} & 94.78\textsubscript{\scriptsize $\pm$0.42} & 55.52\textsubscript{\scriptsize $\pm$6.14} \\
pill & 98.25\textsubscript{\scriptsize $\pm$0.40} & 73.27\textsubscript{\scriptsize $\pm$5.01} & 98.38\textsubscript{\scriptsize $\pm$0.23} & 72.17\textsubscript{\scriptsize $\pm$0.35} & 97.32\textsubscript{\scriptsize $\pm$0.05} & 39.17\textsubscript{\scriptsize $\pm$2.99} \\
screw & 89.81\textsubscript{\scriptsize $\pm$1.92} & 44.54\textsubscript{\scriptsize $\pm$11.54} & 98.53\textsubscript{\scriptsize $\pm$0.39} & 91.66\textsubscript{\scriptsize $\pm$2.43} & 92.52\textsubscript{\scriptsize $\pm$1.51} & 69.55\textsubscript{\scriptsize $\pm$6.73} \\
tile & 100.00\textsubscript{\scriptsize $\pm$0.00} & 99.31\textsubscript{\scriptsize $\pm$0.72} & 95.48\textsubscript{\scriptsize $\pm$0.09} & 90.45\textsubscript{\scriptsize $\pm$2.99} & 86.94\textsubscript{\scriptsize $\pm$0.74} & 68.33\textsubscript{\scriptsize $\pm$5.67} \\
toothbrush & 97.78\textsubscript{\scriptsize $\pm$0.48} & 82.96\textsubscript{\scriptsize $\pm$8.36} & 99.08\textsubscript{\scriptsize $\pm$0.06} & 97.22\textsubscript{\scriptsize $\pm$0.35} & 90.03\textsubscript{\scriptsize $\pm$1.37} & 82.19\textsubscript{\scriptsize $\pm$4.26} \\
transistor & 97.15\textsubscript{\scriptsize $\pm$1.18} & 53.81\textsubscript{\scriptsize $\pm$12.77} & 92.82\textsubscript{\scriptsize $\pm$1.86} & 42.23\textsubscript{\scriptsize $\pm$8.31} & 77.97\textsubscript{\scriptsize $\pm$2.14} & 22.38\textsubscript{\scriptsize $\pm$5.13} \\
wood & 99.06\textsubscript{\scriptsize $\pm$0.41} & 93.57\textsubscript{\scriptsize $\pm$3.29} & 95.44\textsubscript{\scriptsize $\pm$0.88} & 84.95\textsubscript{\scriptsize $\pm$1.02} & 91.58\textsubscript{\scriptsize $\pm$1.18} & 73.34\textsubscript{\scriptsize $\pm$1.30} \\
zipper & 99.12\textsubscript{\scriptsize $\pm$0.41} & 89.99\textsubscript{\scriptsize $\pm$3.19} & 95.79\textsubscript{\scriptsize $\pm$0.38} & 61.22\textsubscript{\scriptsize $\pm$6.61} & 88.56\textsubscript{\scriptsize $\pm$0.79} & 50.87\textsubscript{\scriptsize $\pm$5.76} \\
\midrule
\textbf{Average} & \textbf{98.16\textsubscript{\scriptsize $\pm$0.19}} & \textbf{82.57\textsubscript{\scriptsize $\pm$1.19}} & \textbf{97.58\textsubscript{\scriptsize $\pm$0.11}} & \textbf{83.51\textsubscript{\scriptsize $\pm$0.49}} & \textbf{92.51\textsubscript{\scriptsize $\pm$0.13}} & \textbf{67.91\textsubscript{\scriptsize $\pm$0.69}} \\
\bottomrule
\end{tabular}}
\label{tab:mvtec_8shot}
\end{table}

\begin{table}[H]
\centering
\caption{Per-category results on VisA (1-shot). Each cell reports mean$\pm$std across 3 seeds for clean-defended and PGD-20 ($\epsilon=8/255$) robust evaluation. All values are percentages.}
\setlength{\tabcolsep}{4pt}
\resizebox{\linewidth}{!}{%
\begin{tabular}{lcccccc}
\toprule
 & \multicolumn{2}{c}{Image\ AUROC} & \multicolumn{2}{c}{Pixel\ AUROC} & \multicolumn{2}{c}{PRO} \\
\cmidrule(lr){2-3} \cmidrule(lr){4-5} \cmidrule(lr){6-7}
Object & Clean & Robust & Clean & Robust & Clean & Robust \\
\midrule
candle & 87.05\textsubscript{\scriptsize $\pm$0.06} & 71.90\textsubscript{\scriptsize $\pm$10.61} & 98.38\textsubscript{\scriptsize $\pm$0.82} & 85.66\textsubscript{\scriptsize $\pm$3.43} & 94.80\textsubscript{\scriptsize $\pm$0.10} & 72.31\textsubscript{\scriptsize $\pm$5.46} \\
capsules & 96.96\textsubscript{\scriptsize $\pm$0.90} & 64.54\textsubscript{\scriptsize $\pm$5.93} & 98.06\textsubscript{\scriptsize $\pm$0.20} & 85.72\textsubscript{\scriptsize $\pm$0.46} & 92.36\textsubscript{\scriptsize $\pm$0.29} & 68.80\textsubscript{\scriptsize $\pm$1.65} \\
cashew & 89.19\textsubscript{\scriptsize $\pm$3.52} & 35.95\textsubscript{\scriptsize $\pm$15.88} & 99.56\textsubscript{\scriptsize $\pm$0.03} & 90.65\textsubscript{\scriptsize $\pm$2.76} & 91.57\textsubscript{\scriptsize $\pm$1.70} & 71.15\textsubscript{\scriptsize $\pm$6.02} \\
chewinggum & 97.38\textsubscript{\scriptsize $\pm$0.34} & 66.14\textsubscript{\scriptsize $\pm$17.43} & 99.44\textsubscript{\scriptsize $\pm$0.01} & 95.88\textsubscript{\scriptsize $\pm$1.67} & 89.18\textsubscript{\scriptsize $\pm$1.39} & 66.46\textsubscript{\scriptsize $\pm$8.44} \\
fryum & 88.99\textsubscript{\scriptsize $\pm$1.04} & 54.94\textsubscript{\scriptsize $\pm$6.17} & 97.14\textsubscript{\scriptsize $\pm$0.21} & 83.38\textsubscript{\scriptsize $\pm$2.32} & 87.94\textsubscript{\scriptsize $\pm$0.48} & 71.24\textsubscript{\scriptsize $\pm$3.86} \\
macaroni1 & 74.84\textsubscript{\scriptsize $\pm$5.58} & 33.83\textsubscript{\scriptsize $\pm$9.26} & 98.57\textsubscript{\scriptsize $\pm$0.52} & 84.88\textsubscript{\scriptsize $\pm$3.29} & 93.44\textsubscript{\scriptsize $\pm$2.62} & 52.59\textsubscript{\scriptsize $\pm$7.97} \\
macaroni2 & 52.45\textsubscript{\scriptsize $\pm$1.52} & 33.20\textsubscript{\scriptsize $\pm$8.15} & 95.16\textsubscript{\scriptsize $\pm$0.10} & 83.38\textsubscript{\scriptsize $\pm$3.76} & 78.25\textsubscript{\scriptsize $\pm$2.72} & 56.82\textsubscript{\scriptsize $\pm$6.72} \\
pcb1 & 96.08\textsubscript{\scriptsize $\pm$0.94} & 74.45\textsubscript{\scriptsize $\pm$4.15} & 99.32\textsubscript{\scriptsize $\pm$0.02} & 98.66\textsubscript{\scriptsize $\pm$0.16} & 93.35\textsubscript{\scriptsize $\pm$0.65} & 83.62\textsubscript{\scriptsize $\pm$3.18} \\
pcb2 & 87.60\textsubscript{\scriptsize $\pm$0.47} & 56.07\textsubscript{\scriptsize $\pm$3.69} & 95.86\textsubscript{\scriptsize $\pm$0.16} & 93.03\textsubscript{\scriptsize $\pm$0.70} & 86.17\textsubscript{\scriptsize $\pm$0.69} & 71.85\textsubscript{\scriptsize $\pm$1.92} \\
pcb3 & 79.59\textsubscript{\scriptsize $\pm$5.55} & 45.34\textsubscript{\scriptsize $\pm$12.10} & 96.56\textsubscript{\scriptsize $\pm$0.20} & 94.19\textsubscript{\scriptsize $\pm$0.34} & 85.37\textsubscript{\scriptsize $\pm$0.73} & 69.55\textsubscript{\scriptsize $\pm$1.79} \\
pcb4 & 85.85\textsubscript{\scriptsize $\pm$12.27} & 36.87\textsubscript{\scriptsize $\pm$13.68} & 97.45\textsubscript{\scriptsize $\pm$1.06} & 86.23\textsubscript{\scriptsize $\pm$4.66} & 84.94\textsubscript{\scriptsize $\pm$6.82} & 45.46\textsubscript{\scriptsize $\pm$17.31} \\
pipe\_fryum & 89.36\textsubscript{\scriptsize $\pm$3.42} & 46.11\textsubscript{\scriptsize $\pm$4.72} & 99.30\textsubscript{\scriptsize $\pm$0.08} & 86.39\textsubscript{\scriptsize $\pm$3.65} & 96.22\textsubscript{\scriptsize $\pm$0.60} & 76.66\textsubscript{\scriptsize $\pm$1.71} \\
\midrule
\textbf{Average} & \textbf{85.45\textsubscript{\scriptsize $\pm$1.08}} & \textbf{51.61\textsubscript{\scriptsize $\pm$0.88}} & \textbf{97.90\textsubscript{\scriptsize $\pm$0.11}} & \textbf{89.01\textsubscript{\scriptsize $\pm$0.11}} & \textbf{89.47\textsubscript{\scriptsize $\pm$0.68}} & \textbf{67.21\textsubscript{\scriptsize $\pm$0.55}} \\
\bottomrule
\end{tabular}}
\label{tab:visa_1shot}
\end{table}

\begin{table}[H]
\centering
\caption{Per-category results on VisA (2-shot). Each cell reports mean$\pm$std across 3 seeds for clean-defended and PGD-20 ($\epsilon=8/255$) robust evaluation. All values are percentages.}
\setlength{\tabcolsep}{4pt}
\resizebox{\linewidth}{!}{%
\begin{tabular}{lcccccc}
\toprule
 & \multicolumn{2}{c}{Image\ AUROC} & \multicolumn{2}{c}{Pixel\ AUROC} & \multicolumn{2}{c}{PRO} \\
\cmidrule(lr){2-3} \cmidrule(lr){4-5} \cmidrule(lr){6-7}
Object & Clean & Robust & Clean & Robust & Clean & Robust \\
\midrule
candle & 89.19\textsubscript{\scriptsize $\pm$3.62} & 72.43\textsubscript{\scriptsize $\pm$3.94} & 98.91\textsubscript{\scriptsize $\pm$0.04} & 83.20\textsubscript{\scriptsize $\pm$4.71} & 95.13\textsubscript{\scriptsize $\pm$0.40} & 61.21\textsubscript{\scriptsize $\pm$7.40} \\
capsules & 97.73\textsubscript{\scriptsize $\pm$0.46} & 67.96\textsubscript{\scriptsize $\pm$6.14} & 98.08\textsubscript{\scriptsize $\pm$0.28} & 88.03\textsubscript{\scriptsize $\pm$1.14} & 91.88\textsubscript{\scriptsize $\pm$0.53} & 71.44\textsubscript{\scriptsize $\pm$0.65} \\
cashew & 91.03\textsubscript{\scriptsize $\pm$1.71} & 44.66\textsubscript{\scriptsize $\pm$8.38} & 99.56\textsubscript{\scriptsize $\pm$0.03} & 90.97\textsubscript{\scriptsize $\pm$1.83} & 91.73\textsubscript{\scriptsize $\pm$0.96} & 73.54\textsubscript{\scriptsize $\pm$1.48} \\
chewinggum & 97.94\textsubscript{\scriptsize $\pm$0.51} & 82.18\textsubscript{\scriptsize $\pm$4.58} & 99.40\textsubscript{\scriptsize $\pm$0.02} & 97.42\textsubscript{\scriptsize $\pm$0.62} & 89.87\textsubscript{\scriptsize $\pm$1.27} & 71.56\textsubscript{\scriptsize $\pm$6.30} \\
fryum & 93.06\textsubscript{\scriptsize $\pm$0.81} & 60.59\textsubscript{\scriptsize $\pm$2.79} & 97.47\textsubscript{\scriptsize $\pm$0.19} & 85.35\textsubscript{\scriptsize $\pm$1.96} & 90.09\textsubscript{\scriptsize $\pm$0.52} & 75.95\textsubscript{\scriptsize $\pm$2.42} \\
macaroni1 & 75.71\textsubscript{\scriptsize $\pm$3.25} & 36.84\textsubscript{\scriptsize $\pm$6.13} & 99.00\textsubscript{\scriptsize $\pm$0.19} & 84.25\textsubscript{\scriptsize $\pm$2.00} & 94.37\textsubscript{\scriptsize $\pm$1.28} & 55.09\textsubscript{\scriptsize $\pm$6.20} \\
macaroni2 & 55.70\textsubscript{\scriptsize $\pm$3.34} & 38.42\textsubscript{\scriptsize $\pm$1.21} & 95.55\textsubscript{\scriptsize $\pm$0.17} & 84.19\textsubscript{\scriptsize $\pm$4.41} & 79.69\textsubscript{\scriptsize $\pm$0.82} & 52.23\textsubscript{\scriptsize $\pm$13.40} \\
pcb1 & 96.24\textsubscript{\scriptsize $\pm$0.74} & 72.73\textsubscript{\scriptsize $\pm$6.60} & 99.36\textsubscript{\scriptsize $\pm$0.02} & 98.61\textsubscript{\scriptsize $\pm$0.10} & 93.23\textsubscript{\scriptsize $\pm$0.29} & 82.72\textsubscript{\scriptsize $\pm$2.54} \\
pcb2 & 87.67\textsubscript{\scriptsize $\pm$1.49} & 62.67\textsubscript{\scriptsize $\pm$3.35} & 95.89\textsubscript{\scriptsize $\pm$0.15} & 93.51\textsubscript{\scriptsize $\pm$0.61} & 86.38\textsubscript{\scriptsize $\pm$1.17} & 73.49\textsubscript{\scriptsize $\pm$2.79} \\
pcb3 & 86.71\textsubscript{\scriptsize $\pm$3.36} & 49.45\textsubscript{\scriptsize $\pm$9.52} & 97.16\textsubscript{\scriptsize $\pm$0.15} & 94.69\textsubscript{\scriptsize $\pm$1.35} & 87.24\textsubscript{\scriptsize $\pm$1.00} & 70.16\textsubscript{\scriptsize $\pm$2.63} \\
pcb4 & 93.26\textsubscript{\scriptsize $\pm$6.77} & 49.99\textsubscript{\scriptsize $\pm$10.80} & 98.11\textsubscript{\scriptsize $\pm$0.49} & 89.11\textsubscript{\scriptsize $\pm$1.97} & 87.90\textsubscript{\scriptsize $\pm$4.28} & 59.43\textsubscript{\scriptsize $\pm$8.39} \\
pipe\_fryum & 90.93\textsubscript{\scriptsize $\pm$3.54} & 51.33\textsubscript{\scriptsize $\pm$4.14} & 99.38\textsubscript{\scriptsize $\pm$0.12} & 91.06\textsubscript{\scriptsize $\pm$1.62} & 96.10\textsubscript{\scriptsize $\pm$0.45} & 78.13\textsubscript{\scriptsize $\pm$3.42} \\
\midrule
\textbf{Average} & \textbf{87.93\textsubscript{\scriptsize $\pm$1.44}} & \textbf{57.44\textsubscript{\scriptsize $\pm$1.18}} & \textbf{98.16\textsubscript{\scriptsize $\pm$0.04}} & \textbf{90.03\textsubscript{\scriptsize $\pm$0.49}} & \textbf{90.30\textsubscript{\scriptsize $\pm$0.48}} & \textbf{68.75\textsubscript{\scriptsize $\pm$0.71}} \\
\bottomrule
\end{tabular}}
\label{tab:visa_2shot}
\end{table}

\begin{table}[H]
\centering
\caption{Per-category results on VisA (4-shot). Each cell reports mean$\pm$std across 3 seeds for clean-defended and PGD-20 ($\epsilon=8/255$) robust evaluation. All values are percentages.}
\setlength{\tabcolsep}{4pt}
\resizebox{\linewidth}{!}{%
\begin{tabular}{lcccccc}
\toprule
 & \multicolumn{2}{c}{Image\ AUROC} & \multicolumn{2}{c}{Pixel\ AUROC} & \multicolumn{2}{c}{PRO} \\
\cmidrule(lr){2-3} \cmidrule(lr){4-5} \cmidrule(lr){6-7}
Object & Clean & Robust & Clean & Robust & Clean & Robust \\
\midrule
candle & 90.67\textsubscript{\scriptsize $\pm$3.48} & 73.86\textsubscript{\scriptsize $\pm$6.08} & 98.83\textsubscript{\scriptsize $\pm$0.37} & 86.72\textsubscript{\scriptsize $\pm$1.05} & 95.29\textsubscript{\scriptsize $\pm$0.18} & 75.17\textsubscript{\scriptsize $\pm$1.10} \\
capsules & 97.83\textsubscript{\scriptsize $\pm$0.18} & 74.00\textsubscript{\scriptsize $\pm$4.25} & 97.67\textsubscript{\scriptsize $\pm$1.06} & 90.13\textsubscript{\scriptsize $\pm$0.31} & 93.10\textsubscript{\scriptsize $\pm$0.66} & 76.14\textsubscript{\scriptsize $\pm$1.42} \\
cashew & 93.37\textsubscript{\scriptsize $\pm$0.62} & 43.48\textsubscript{\scriptsize $\pm$12.84} & 99.60\textsubscript{\scriptsize $\pm$0.01} & 90.42\textsubscript{\scriptsize $\pm$1.68} & 92.75\textsubscript{\scriptsize $\pm$0.64} & 72.84\textsubscript{\scriptsize $\pm$1.42} \\
chewinggum & 98.01\textsubscript{\scriptsize $\pm$0.48} & 88.57\textsubscript{\scriptsize $\pm$3.13} & 99.43\textsubscript{\scriptsize $\pm$0.01} & 97.88\textsubscript{\scriptsize $\pm$0.36} & 90.04\textsubscript{\scriptsize $\pm$0.55} & 69.76\textsubscript{\scriptsize $\pm$8.66} \\
fryum & 93.47\textsubscript{\scriptsize $\pm$1.32} & 57.68\textsubscript{\scriptsize $\pm$1.17} & 97.75\textsubscript{\scriptsize $\pm$0.10} & 85.40\textsubscript{\scriptsize $\pm$0.68} & 90.89\textsubscript{\scriptsize $\pm$0.32} & 77.05\textsubscript{\scriptsize $\pm$0.44} \\
macaroni1 & 81.46\textsubscript{\scriptsize $\pm$6.29} & 46.93\textsubscript{\scriptsize $\pm$1.11} & 99.38\textsubscript{\scriptsize $\pm$0.26} & 90.22\textsubscript{\scriptsize $\pm$2.31} & 96.38\textsubscript{\scriptsize $\pm$1.24} & 70.41\textsubscript{\scriptsize $\pm$6.52} \\
macaroni2 & 56.32\textsubscript{\scriptsize $\pm$1.53} & 38.91\textsubscript{\scriptsize $\pm$3.49} & 96.08\textsubscript{\scriptsize $\pm$0.43} & 85.37\textsubscript{\scriptsize $\pm$1.77} & 81.59\textsubscript{\scriptsize $\pm$2.20} & 50.98\textsubscript{\scriptsize $\pm$3.72} \\
pcb1 & 97.71\textsubscript{\scriptsize $\pm$1.10} & 78.06\textsubscript{\scriptsize $\pm$3.43} & 99.38\textsubscript{\scriptsize $\pm$0.01} & 98.72\textsubscript{\scriptsize $\pm$0.09} & 93.46\textsubscript{\scriptsize $\pm$0.24} & 84.33\textsubscript{\scriptsize $\pm$1.16} \\
pcb2 & 88.65\textsubscript{\scriptsize $\pm$0.50} & 64.58\textsubscript{\scriptsize $\pm$1.98} & 96.13\textsubscript{\scriptsize $\pm$0.08} & 93.70\textsubscript{\scriptsize $\pm$0.09} & 87.29\textsubscript{\scriptsize $\pm$0.18} & 73.42\textsubscript{\scriptsize $\pm$0.22} \\
pcb3 & 87.49\textsubscript{\scriptsize $\pm$2.29} & 52.32\textsubscript{\scriptsize $\pm$9.85} & 97.19\textsubscript{\scriptsize $\pm$0.20} & 95.06\textsubscript{\scriptsize $\pm$0.76} & 87.84\textsubscript{\scriptsize $\pm$1.05} & 73.81\textsubscript{\scriptsize $\pm$0.32} \\
pcb4 & 98.05\textsubscript{\scriptsize $\pm$1.43} & 54.80\textsubscript{\scriptsize $\pm$5.91} & 98.58\textsubscript{\scriptsize $\pm$0.05} & 90.69\textsubscript{\scriptsize $\pm$0.69} & 90.92\textsubscript{\scriptsize $\pm$0.91} & 63.84\textsubscript{\scriptsize $\pm$2.28} \\
pipe\_fryum & 90.99\textsubscript{\scriptsize $\pm$2.66} & 68.33\textsubscript{\scriptsize $\pm$2.10} & 99.41\textsubscript{\scriptsize $\pm$0.07} & 90.56\textsubscript{\scriptsize $\pm$1.56} & 95.93\textsubscript{\scriptsize $\pm$0.12} & 77.51\textsubscript{\scriptsize $\pm$3.90} \\
\midrule
\textbf{Average} & \textbf{89.50\textsubscript{\scriptsize $\pm$1.04}} & \textbf{61.79\textsubscript{\scriptsize $\pm$0.96}} & \textbf{98.29\textsubscript{\scriptsize $\pm$0.07}} & \textbf{91.24\textsubscript{\scriptsize $\pm$0.15}} & \textbf{91.29\textsubscript{\scriptsize $\pm$0.12}} & \textbf{72.11\textsubscript{\scriptsize $\pm$1.16}} \\
\bottomrule
\end{tabular}}
\label{tab:visa_4shot}
\end{table}

\begin{table}[H]
\centering
\caption{Per-category results on VisA (8-shot). Each cell reports mean$\pm$std across 3 seeds for clean-defended and PGD-20 ($\epsilon=8/255$) robust evaluation. All values are percentages.}
\setlength{\tabcolsep}{4pt}
\resizebox{\linewidth}{!}{%
\begin{tabular}{lcccccc}
\toprule
 & \multicolumn{2}{c}{Image\ AUROC} & \multicolumn{2}{c}{Pixel\ AUROC} & \multicolumn{2}{c}{PRO} \\
\cmidrule(lr){2-3} \cmidrule(lr){4-5} \cmidrule(lr){6-7}
Object & Clean & Robust & Clean & Robust & Clean & Robust \\
\midrule
candle & 92.78\textsubscript{\scriptsize $\pm$2.17} & 73.17\textsubscript{\scriptsize $\pm$3.12} & 98.41\textsubscript{\scriptsize $\pm$0.43} & 86.52\textsubscript{\scriptsize $\pm$1.80} & 95.73\textsubscript{\scriptsize $\pm$0.23} & 76.72\textsubscript{\scriptsize $\pm$1.49} \\
capsules & 97.68\textsubscript{\scriptsize $\pm$0.23} & 75.18\textsubscript{\scriptsize $\pm$1.49} & 96.78\textsubscript{\scriptsize $\pm$0.19} & 90.25\textsubscript{\scriptsize $\pm$1.19} & 93.10\textsubscript{\scriptsize $\pm$0.59} & 79.51\textsubscript{\scriptsize $\pm$2.59} \\
cashew & 94.07\textsubscript{\scriptsize $\pm$0.90} & 48.36\textsubscript{\scriptsize $\pm$15.87} & 99.59\textsubscript{\scriptsize $\pm$0.02} & 91.49\textsubscript{\scriptsize $\pm$0.43} & 92.36\textsubscript{\scriptsize $\pm$0.59} & 72.44\textsubscript{\scriptsize $\pm$1.90} \\
chewinggum & 98.25\textsubscript{\scriptsize $\pm$0.51} & 88.63\textsubscript{\scriptsize $\pm$2.78} & 99.42\textsubscript{\scriptsize $\pm$0.02} & 98.19\textsubscript{\scriptsize $\pm$0.50} & 90.23\textsubscript{\scriptsize $\pm$0.35} & 73.16\textsubscript{\scriptsize $\pm$3.53} \\
fryum & 94.71\textsubscript{\scriptsize $\pm$0.54} & 68.05\textsubscript{\scriptsize $\pm$9.03} & 97.82\textsubscript{\scriptsize $\pm$0.09} & 86.15\textsubscript{\scriptsize $\pm$1.56} & 90.95\textsubscript{\scriptsize $\pm$0.16} & 78.28\textsubscript{\scriptsize $\pm$2.28} \\
macaroni1 & 80.87\textsubscript{\scriptsize $\pm$2.39} & 34.79\textsubscript{\scriptsize $\pm$2.13} & 99.26\textsubscript{\scriptsize $\pm$0.17} & 86.96\textsubscript{\scriptsize $\pm$1.78} & 96.24\textsubscript{\scriptsize $\pm$0.66} & 61.94\textsubscript{\scriptsize $\pm$4.36} \\
macaroni2 & 59.46\textsubscript{\scriptsize $\pm$1.81} & 36.71\textsubscript{\scriptsize $\pm$1.93} & 96.51\textsubscript{\scriptsize $\pm$0.34} & 85.03\textsubscript{\scriptsize $\pm$1.81} & 85.22\textsubscript{\scriptsize $\pm$1.63} & 53.78\textsubscript{\scriptsize $\pm$6.51} \\
pcb1 & 97.38\textsubscript{\scriptsize $\pm$0.19} & 74.53\textsubscript{\scriptsize $\pm$2.80} & 99.41\textsubscript{\scriptsize $\pm$0.02} & 98.68\textsubscript{\scriptsize $\pm$0.04} & 94.32\textsubscript{\scriptsize $\pm$0.35} & 82.99\textsubscript{\scriptsize $\pm$0.35} \\
pcb2 & 88.44\textsubscript{\scriptsize $\pm$0.14} & 62.80\textsubscript{\scriptsize $\pm$5.13} & 96.26\textsubscript{\scriptsize $\pm$0.07} & 94.02\textsubscript{\scriptsize $\pm$0.23} & 88.72\textsubscript{\scriptsize $\pm$0.34} & 75.38\textsubscript{\scriptsize $\pm$1.75} \\
pcb3 & 91.23\textsubscript{\scriptsize $\pm$1.28} & 54.11\textsubscript{\scriptsize $\pm$6.59} & 97.31\textsubscript{\scriptsize $\pm$0.03} & 94.84\textsubscript{\scriptsize $\pm$0.41} & 89.49\textsubscript{\scriptsize $\pm$0.80} & 75.36\textsubscript{\scriptsize $\pm$1.45} \\
pcb4 & 98.90\textsubscript{\scriptsize $\pm$0.39} & 61.38\textsubscript{\scriptsize $\pm$13.19} & 98.84\textsubscript{\scriptsize $\pm$0.05} & 91.96\textsubscript{\scriptsize $\pm$1.89} & 92.60\textsubscript{\scriptsize $\pm$0.84} & 66.27\textsubscript{\scriptsize $\pm$7.60} \\
pipe\_fryum & 92.73\textsubscript{\scriptsize $\pm$2.03} & 66.18\textsubscript{\scriptsize $\pm$2.09} & 99.44\textsubscript{\scriptsize $\pm$0.05} & 93.00\textsubscript{\scriptsize $\pm$1.17} & 95.87\textsubscript{\scriptsize $\pm$0.25} & 83.96\textsubscript{\scriptsize $\pm$2.98} \\
\midrule
\textbf{Average} & \textbf{90.54\textsubscript{\scriptsize $\pm$0.29}} & \textbf{61.99\textsubscript{\scriptsize $\pm$2.03}} & \textbf{98.26\textsubscript{\scriptsize $\pm$0.03}} & \textbf{91.43\textsubscript{\scriptsize $\pm$0.51}} & \textbf{92.07\textsubscript{\scriptsize $\pm$0.09}} & \textbf{73.32\textsubscript{\scriptsize $\pm$0.40}} \\
\bottomrule
\end{tabular}}
\label{tab:visa_8shot}
\end{table}

\begin{table}[H]
\centering
\caption{Per-category results on MPDD (1-shot). Each cell reports mean$\pm$std across 3 seeds for clean-defended and PGD-20 ($\epsilon=8/255$) robust evaluation. All values are percentages.}
\setlength{\tabcolsep}{4pt}
\resizebox{\linewidth}{!}{%
\begin{tabular}{lcccccc}
\toprule
 & \multicolumn{2}{c}{Image\ AUROC} & \multicolumn{2}{c}{Pixel\ AUROC} & \multicolumn{2}{c}{PRO} \\
\cmidrule(lr){2-3} \cmidrule(lr){4-5} \cmidrule(lr){6-7}
Object & Clean & Robust & Clean & Robust & Clean & Robust \\
\midrule
bracket\_black & 42.71\textsubscript{\scriptsize $\pm$1.91} & 35.20\textsubscript{\scriptsize $\pm$3.16} & 86.98\textsubscript{\scriptsize $\pm$3.17} & 78.74\textsubscript{\scriptsize $\pm$4.55} & 78.43\textsubscript{\scriptsize $\pm$2.81} & 65.76\textsubscript{\scriptsize $\pm$6.13} \\
bracket\_brown & 50.43\textsubscript{\scriptsize $\pm$6.60} & 44.07\textsubscript{\scriptsize $\pm$7.69} & 90.85\textsubscript{\scriptsize $\pm$3.05} & 55.71\textsubscript{\scriptsize $\pm$3.94} & 77.17\textsubscript{\scriptsize $\pm$3.41} & 29.53\textsubscript{\scriptsize $\pm$4.62} \\
bracket\_white & 79.19\textsubscript{\scriptsize $\pm$4.90} & 47.26\textsubscript{\scriptsize $\pm$9.92} & 92.28\textsubscript{\scriptsize $\pm$0.25} & 43.29\textsubscript{\scriptsize $\pm$5.39} & 90.40\textsubscript{\scriptsize $\pm$0.61} & 15.68\textsubscript{\scriptsize $\pm$5.32} \\
connector & 69.13\textsubscript{\scriptsize $\pm$1.45} & 36.67\textsubscript{\scriptsize $\pm$1.67} & 96.75\textsubscript{\scriptsize $\pm$0.23} & 79.36\textsubscript{\scriptsize $\pm$3.34} & 88.78\textsubscript{\scriptsize $\pm$0.77} & 31.95\textsubscript{\scriptsize $\pm$9.90} \\
metal\_plate & 100.00\textsubscript{\scriptsize $\pm$0.00} & 78.08\textsubscript{\scriptsize $\pm$4.58} & 98.59\textsubscript{\scriptsize $\pm$0.15} & 87.94\textsubscript{\scriptsize $\pm$0.60} & 93.17\textsubscript{\scriptsize $\pm$0.15} & 66.17\textsubscript{\scriptsize $\pm$1.42} \\
tubes & 92.18\textsubscript{\scriptsize $\pm$2.19} & 71.17\textsubscript{\scriptsize $\pm$8.50} & 98.64\textsubscript{\scriptsize $\pm$0.17} & 94.80\textsubscript{\scriptsize $\pm$1.01} & 94.37\textsubscript{\scriptsize $\pm$0.46} & 80.56\textsubscript{\scriptsize $\pm$3.68} \\
\midrule
\textbf{Average} & \textbf{72.27\textsubscript{\scriptsize $\pm$2.38}} & \textbf{52.07\textsubscript{\scriptsize $\pm$2.17}} & \textbf{94.02\textsubscript{\scriptsize $\pm$1.00}} & \textbf{73.31\textsubscript{\scriptsize $\pm$0.96}} & \textbf{87.05\textsubscript{\scriptsize $\pm$0.71}} & \textbf{48.27\textsubscript{\scriptsize $\pm$2.91}} \\
\bottomrule
\end{tabular}}
\label{tab:mpdd_1shot}
\end{table}

\begin{table}[h]
\centering
\caption{Per-category results on MPDD (2-shot). Each cell reports mean$\pm$std across 3 seeds for clean-defended and PGD-20 ($\epsilon=8/255$) robust evaluation. All values are percentages.}
\setlength{\tabcolsep}{4pt}
\resizebox{\linewidth}{!}{%
\begin{tabular}{lcccccc}
\toprule
 & \multicolumn{2}{c}{Image\ AUROC} & \multicolumn{2}{c}{Pixel\ AUROC} & \multicolumn{2}{c}{PRO} \\
\cmidrule(lr){2-3} \cmidrule(lr){4-5} \cmidrule(lr){6-7}
Object & Clean & Robust & Clean & Robust & Clean & Robust \\
\midrule
bracket\_black & 48.32\textsubscript{\scriptsize $\pm$12.97} & 33.20\textsubscript{\scriptsize $\pm$14.49} & 88.64\textsubscript{\scriptsize $\pm$1.93} & 76.66\textsubscript{\scriptsize $\pm$6.89} & 81.13\textsubscript{\scriptsize $\pm$2.56} & 63.11\textsubscript{\scriptsize $\pm$5.67} \\
bracket\_brown & 53.95\textsubscript{\scriptsize $\pm$4.64} & 39.47\textsubscript{\scriptsize $\pm$10.21} & 93.48\textsubscript{\scriptsize $\pm$1.25} & 35.74\textsubscript{\scriptsize $\pm$11.10} & 80.78\textsubscript{\scriptsize $\pm$3.29} & 15.69\textsubscript{\scriptsize $\pm$9.04} \\
bracket\_white & 78.15\textsubscript{\scriptsize $\pm$6.86} & 58.96\textsubscript{\scriptsize $\pm$7.51} & 93.04\textsubscript{\scriptsize $\pm$2.99} & 71.99\textsubscript{\scriptsize $\pm$23.08} & 82.09\textsubscript{\scriptsize $\pm$9.83} & 55.40\textsubscript{\scriptsize $\pm$33.97} \\
connector & 73.97\textsubscript{\scriptsize $\pm$6.35} & 41.83\textsubscript{\scriptsize $\pm$12.59} & 97.35\textsubscript{\scriptsize $\pm$0.63} & 83.86\textsubscript{\scriptsize $\pm$5.17} & 90.77\textsubscript{\scriptsize $\pm$2.11} & 45.37\textsubscript{\scriptsize $\pm$16.24} \\
metal\_plate & 100.00\textsubscript{\scriptsize $\pm$0.00} & 74.68\textsubscript{\scriptsize $\pm$2.63} & 98.72\textsubscript{\scriptsize $\pm$0.03} & 88.62\textsubscript{\scriptsize $\pm$0.76} & 93.47\textsubscript{\scriptsize $\pm$0.02} & 66.93\textsubscript{\scriptsize $\pm$1.77} \\
tubes & 91.50\textsubscript{\scriptsize $\pm$2.02} & 66.58\textsubscript{\scriptsize $\pm$6.86} & 98.78\textsubscript{\scriptsize $\pm$0.17} & 95.07\textsubscript{\scriptsize $\pm$0.37} & 94.98\textsubscript{\scriptsize $\pm$0.73} & 81.15\textsubscript{\scriptsize $\pm$1.48} \\
\midrule
\textbf{Average} & \textbf{74.31\textsubscript{\scriptsize $\pm$4.52}} & \textbf{52.45\textsubscript{\scriptsize $\pm$5.92}} & \textbf{95.00\textsubscript{\scriptsize $\pm$0.27}} & \textbf{75.32\textsubscript{\scriptsize $\pm$4.14}} & \textbf{87.20\textsubscript{\scriptsize $\pm$1.90}} & \textbf{54.61\textsubscript{\scriptsize $\pm$6.99}} \\
\bottomrule
\end{tabular}}
\label{tab:mpdd_2shot}
\end{table}

\begin{table}[h]
\centering
\caption{Per-category results on MPDD (4-shot). Each cell reports mean$\pm$std across 3 seeds for clean-defended and PGD-20 ($\epsilon=8/255$) robust evaluation. All values are percentages.}
\setlength{\tabcolsep}{4pt}
\resizebox{\linewidth}{!}{%
\begin{tabular}{lcccccc}
\toprule
 & \multicolumn{2}{c}{Image\ AUROC} & \multicolumn{2}{c}{Pixel\ AUROC} & \multicolumn{2}{c}{PRO} \\
\cmidrule(lr){2-3} \cmidrule(lr){4-5} \cmidrule(lr){6-7}
Object & Clean & Robust & Clean & Robust & Clean & Robust \\
\midrule
bracket\_black & 56.65\textsubscript{\scriptsize $\pm$7.95} & 47.89\textsubscript{\scriptsize $\pm$16.16} & 90.35\textsubscript{\scriptsize $\pm$4.82} & 81.36\textsubscript{\scriptsize $\pm$4.46} & 83.39\textsubscript{\scriptsize $\pm$4.18} & 69.82\textsubscript{\scriptsize $\pm$10.60} \\
bracket\_brown & 59.35\textsubscript{\scriptsize $\pm$1.92} & 54.00\textsubscript{\scriptsize $\pm$6.52} & 92.96\textsubscript{\scriptsize $\pm$0.22} & 37.55\textsubscript{\scriptsize $\pm$9.93} & 80.66\textsubscript{\scriptsize $\pm$2.01} & 17.81\textsubscript{\scriptsize $\pm$6.93} \\
bracket\_white & 66.15\textsubscript{\scriptsize $\pm$17.93} & 41.78\textsubscript{\scriptsize $\pm$8.06} & 84.57\textsubscript{\scriptsize $\pm$8.02} & 68.22\textsubscript{\scriptsize $\pm$20.28} & 61.36\textsubscript{\scriptsize $\pm$15.15} & 46.59\textsubscript{\scriptsize $\pm$23.29} \\
connector & 82.14\textsubscript{\scriptsize $\pm$7.08} & 52.86\textsubscript{\scriptsize $\pm$12.84} & 97.76\textsubscript{\scriptsize $\pm$0.88} & 89.78\textsubscript{\scriptsize $\pm$4.38} & 92.16\textsubscript{\scriptsize $\pm$2.97} & 64.66\textsubscript{\scriptsize $\pm$14.20} \\
metal\_plate & 100.00\textsubscript{\scriptsize $\pm$0.00} & 84.13\textsubscript{\scriptsize $\pm$3.98} & 98.78\textsubscript{\scriptsize $\pm$0.10} & 89.96\textsubscript{\scriptsize $\pm$1.62} & 93.54\textsubscript{\scriptsize $\pm$0.24} & 71.38\textsubscript{\scriptsize $\pm$2.98} \\
tubes & 93.72\textsubscript{\scriptsize $\pm$1.69} & 75.92\textsubscript{\scriptsize $\pm$4.30} & 98.89\textsubscript{\scriptsize $\pm$0.14} & 96.16\textsubscript{\scriptsize $\pm$0.49} & 95.22\textsubscript{\scriptsize $\pm$0.58} & 84.75\textsubscript{\scriptsize $\pm$1.78} \\
\midrule
\textbf{Average} & \textbf{76.34\textsubscript{\scriptsize $\pm$3.89}} & \textbf{59.43\textsubscript{\scriptsize $\pm$4.14}} & \textbf{93.89\textsubscript{\scriptsize $\pm$2.00}} & \textbf{77.17\textsubscript{\scriptsize $\pm$2.90}} & \textbf{84.39\textsubscript{\scriptsize $\pm$3.57}} & \textbf{59.17\textsubscript{\scriptsize $\pm$6.44}} \\
\bottomrule
\end{tabular}}
\label{tab:mpdd_4shot}
\end{table}

\begin{table}[H]
\centering
\caption{Per-category results on MPDD (8-shot). Each cell reports mean$\pm$std across 3 seeds for clean-defended and PGD-20 ($\epsilon=8/255$) robust evaluation. All values are percentages.}
\setlength{\tabcolsep}{4pt}
\resizebox{\linewidth}{!}{%
\begin{tabular}{lcccccc}
\toprule
 & \multicolumn{2}{c}{Image\ AUROC} & \multicolumn{2}{c}{Pixel\ AUROC} & \multicolumn{2}{c}{PRO} \\
\cmidrule(lr){2-3} \cmidrule(lr){4-5} \cmidrule(lr){6-7}
Object & Clean & Robust & Clean & Robust & Clean & Robust \\
\midrule
bracket\_black & 60.86\textsubscript{\scriptsize $\pm$0.50} & 46.94\textsubscript{\scriptsize $\pm$12.73} & 93.60\textsubscript{\scriptsize $\pm$2.01} & 86.97\textsubscript{\scriptsize $\pm$5.87} & 86.29\textsubscript{\scriptsize $\pm$2.47} & 75.63\textsubscript{\scriptsize $\pm$7.07} \\
bracket\_brown & 61.36\textsubscript{\scriptsize $\pm$1.93} & 53.65\textsubscript{\scriptsize $\pm$5.92} & 94.08\textsubscript{\scriptsize $\pm$0.40} & 44.62\textsubscript{\scriptsize $\pm$12.91} & 81.95\textsubscript{\scriptsize $\pm$1.60} & 24.75\textsubscript{\scriptsize $\pm$12.80} \\
bracket\_white & 72.63\textsubscript{\scriptsize $\pm$11.30} & 52.37\textsubscript{\scriptsize $\pm$17.42} & 89.17\textsubscript{\scriptsize $\pm$8.30} & 59.95\textsubscript{\scriptsize $\pm$24.82} & 68.20\textsubscript{\scriptsize $\pm$14.15} & 49.02\textsubscript{\scriptsize $\pm$37.55} \\
connector & 85.00\textsubscript{\scriptsize $\pm$3.11} & 53.41\textsubscript{\scriptsize $\pm$4.29} & 98.07\textsubscript{\scriptsize $\pm$0.25} & 91.31\textsubscript{\scriptsize $\pm$0.38} & 93.20\textsubscript{\scriptsize $\pm$0.89} & 70.03\textsubscript{\scriptsize $\pm$1.46} \\
metal\_plate & 100.00\textsubscript{\scriptsize $\pm$0.00} & 87.92\textsubscript{\scriptsize $\pm$10.08} & 98.85\textsubscript{\scriptsize $\pm$0.08} & 91.23\textsubscript{\scriptsize $\pm$2.54} & 93.78\textsubscript{\scriptsize $\pm$0.21} & 74.36\textsubscript{\scriptsize $\pm$6.71} \\
tubes & 93.19\textsubscript{\scriptsize $\pm$0.92} & 70.71\textsubscript{\scriptsize $\pm$7.85} & 98.85\textsubscript{\scriptsize $\pm$0.08} & 95.67\textsubscript{\scriptsize $\pm$0.32} & 95.15\textsubscript{\scriptsize $\pm$0.43} & 83.38\textsubscript{\scriptsize $\pm$1.05} \\
\midrule
\textbf{Average} & \textbf{78.84\textsubscript{\scriptsize $\pm$1.37}} & \textbf{60.83\textsubscript{\scriptsize $\pm$4.37}} & \textbf{95.44\textsubscript{\scriptsize $\pm$1.61}} & \textbf{78.29\textsubscript{\scriptsize $\pm$2.54}} & \textbf{86.43\textsubscript{\scriptsize $\pm$2.88}} & \textbf{62.86\textsubscript{\scriptsize $\pm$4.63}} \\
\bottomrule
\end{tabular}}
\label{tab:mpdd_8shot}
\end{table}

\section{Inference Overhead}

\begin{figure}[h]
  \centering
  \includegraphics[width=0.7\linewidth]{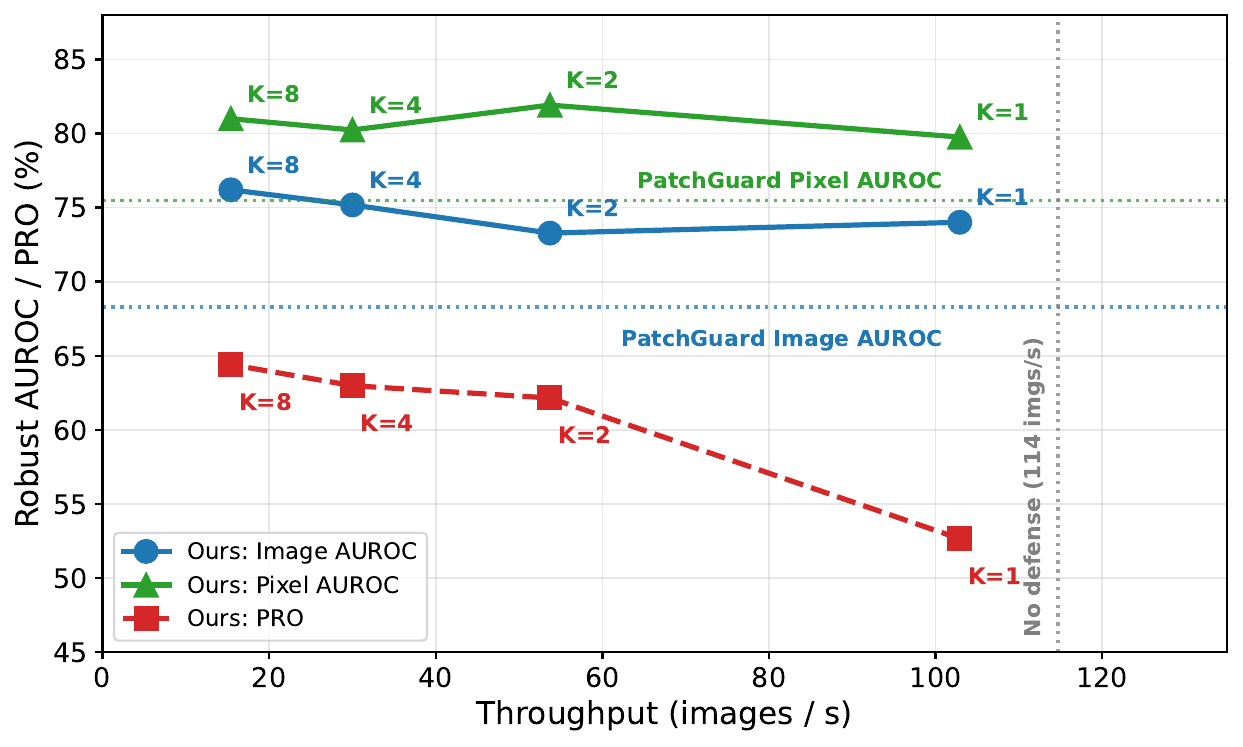}
   \caption{Latency--robustness Pareto on MVTec-AD, $k{=}4$, PGD-20 at $\epsilon{=}8/255$.}
   \label{latency_pareto}
\end{figure}

We sweep $K \in \{1, 2, 4, 8\}$ on MVTec-AD at $k=4$ shots in Figure~\ref{latency_pareto}. Practitioners can trade a small
robustness margin for substantial speed. At $K=1$,
throughput reaches $102.9$ imgs/s ($1.2\times$ slowdown, nearly
matching the undefended detector), and image and pixel AUROC remain
above PatchGuard. Only PRO drops meaningfully, confirming our
ablation finding that PS's median aggregation primarily aids
localization and the FP carries most of image- and
pixel-level detection. The $K$ backbone passes are also parallel across shifts and can be batched into a single forward pass, which we did not exploit in this measurement and which would further reduce wall-clock latency.

\bibliography{egbib}